\documentclass[11pt]{article}

\usepackage{acl}

\usepackage{times}
\usepackage{latexsym}
\usepackage{float}
\usepackage[T1]{fontenc}

\usepackage[utf8]{inputenc}

\usepackage{microtype}
\usepackage{paralist}
\usepackage{inconsolata}

\usepackage{graphicx}

\usepackage{amsmath}
\usepackage{algorithm}
\usepackage{algpseudocode}
\usepackage{xcolor}
\usepackage{booktabs}
\usepackage{longtable}
\usepackage{multirow}
\usepackage{array}
\usepackage{tabularx}
\usepackage{enumitem}
\usepackage{tcolorbox}
\usepackage{adjustbox}
\usepackage{bbm}

\title{\textsc{EpiQAL}: Benchmarking Large Language Models in Epidemiological Question Answering and Reasoning}

\author{
  \textbf{Mingyang Wei\textsuperscript{1}},
  \textbf{Dehai Min\textsuperscript{2}},
  \textbf{Zewen Liu\textsuperscript{1}},
  \textbf{Yuzhang Xie\textsuperscript{1}},
  \textbf{Guanchen Wu\textsuperscript{1}},
  \textbf{Ziyang Zhang\textsuperscript{1}},
  \\
  \textbf{Carl Yang\textsuperscript{1}},
  \textbf{Max S.Y. Lau\textsuperscript{1}},
  \textbf{Qi He\textsuperscript{3}},
  \textbf{Lu Cheng\textsuperscript{2}},
  \textbf{Wei Jin\textsuperscript{1}\thanks{Correspondence: \href{mailto:wei.jin@emory.edu}{wei.jin@emory.edu}}}
  \\
  \\
  \textsuperscript{1}Emory University,\quad
  \textsuperscript{2}University of Illinois Chicago,\quad
  \textsuperscript{3}Microsoft
  \\
  \texttt{\{mingyang.wei, zewen.liu, yuzhang.xie, guanchen.wu, ziyang.zhang2\}@emory.edu}
  \\
  \texttt{\{j.carlyang, msy.lau, wei.jin\}@emory.edu}
  \\
  \texttt{\{dmin10, lucheng\}@uic.edu},\quad
  \texttt{qhe@microsoft.com}
}

\begin{document}
\maketitle
\begin{abstract}
Reliable epidemiological reasoning requires synthesizing study evidence to infer disease burden, transmission dynamics, and intervention effects at the population level. Existing medical question answering benchmarks primarily emphasize clinical knowledge or patient-level reasoning, yet few systematically evaluate evidence-grounded epidemiological inference. We present EpiQAL, to our knowledge the first diagnostic benchmark for epidemiological question answering over research literature, comprising three subsets built from open-access articles across diverse diseases. The three subsets progressively test factual recall, multi-step inference, and conclusion reconstruction under incomplete information, and are constructed through a quality-controlled pipeline combining taxonomy guidance, multi-model verification, and difficulty screening.

Experiments on fifteen models spanning open-source and proprietary systems reveal that current LLMs show limited performance on epidemiological reasoning, with multi-step inference posing the greatest challenge. Model rankings shift across subsets, and scale alone does not predict success. Chain-of-Thought prompting benefits multi-step inference but yields mixed results elsewhere. EpiQAL provides fine-grained diagnostic signals for evidence-grounding, inferential reasoning, and conclusion reconstruction.\footnote{Benchmark and code are available at \url{https://github.com/myweiii/EpiQAL}.}
\end{abstract}

\section{Introduction}
The COVID-19 pandemic underscored the challenge of extracting reliable insights from a rapidly expanding epidemiological literature \citep{WANG20214, dieguezcampa2021_researchpandemic}. Evidence-informed public health practice targets populations rather than individual patients and often demands synthesizing heterogeneous, context-dependent study findings \citep{brownson2009_ebph, pub.1043917632}. Biomedical QA systems retrieve and summarize evidence from large article collections \citep{bauer2012_biomedicalqa, tsatsaronis2015_bioasq, wallace2019_evidence}, and recent agentic AI systems further extend healthcare AI with planning, tool use, and multi-agent collaboration \citep{xu2025agentic}, but these efforts primarily support clinical knowledge retrieval and patient-level decision making. Epidemiological reasoning, by contrast, requires population-level statistical and causal inference about disease burden, transmission dynamics, and intervention effects \citep{glass2013_causal_public_health}. This gap motivates QA benchmarks tailored to epidemiological inference.

A suitable benchmark should be \emph{controlled}, limiting shortcut cues that allow models to exploit superficial patterns such as lexical overlap between questions and contexts \citep{shinoda-etal-2021-question}, and \emph{trustworthy}, anchoring answers to verifiable study evidence rather than relying solely on annotator judgment. Existing resources only partially meet these requirements. Exam-style benchmarks primarily test clinical knowledge \citep{jin2021disease, pmlr-v174-pal22a}; literature-grounded datasets rely on abstracts and constrained label spaces \citep{jin-etal-2019-pubmedqa}; epidemic-focused datasets are typically disease-specific with extractive formats \citep{moller-etal-2020-covid, CoQUADraza, epicqa}. Moreover, expert annotation remains costly, limiting both scale and topic coverage.

We present \textbf{EpiQAL} (\underline{Epi}demiological \underline{QA} over the \underline{L}iterature), which to our knowledge is the first benchmark to systematically evaluate epidemiological QA by combining broad topic coverage, set-based evaluation that accommodates both single- and multi-answer questions, and document-grounded answer derivation. Building EpiQAL requires addressing four challenges.

\begin{compactenum}[(1)]
    \item \textbf{Scope.} A benchmark limited to a single disease cannot assess generalization across the diverse phenomena epidemiological research spans.
    
    \item \textbf{Grounding.} Epidemiological conclusions must be traceable to study evidence. Without such grounding, it is difficult to distinguish genuine inference from hallucination.
    
    \item \textbf{Verification.} Epidemiological questions may admit multiple valid answers, for instance when a study identifies several vector species, all of which are correct responses. Validating such multi-answer correctness at scale without exhaustive expert annotation requires automated quality control.
    
    \item \textbf{Difficulty.} Models can exploit superficial cues such as lexical overlap between question stems and correct options, succeeding without genuine comprehension.
\end{compactenum}

Our framework addresses each challenge. For \emph{scope}, we develop a taxonomy of six categories and twenty-five topics with an epidemiology expert. For \emph{grounding}, we adopt subset-specific strategies anchored in document evidence. For \emph{verification}, we design a checking model group with human review for uncertain cases. For \emph{difficulty}, we employ difficulty screening and stem refinement \citep{bai2024longbench2, wu2025webdancer}. These components yield three subsets: \textbf{EpiQAL-A} tests text-grounded factual recall, \textbf{EpiQAL-B} tests multi-step inference requiring integration of multiple findings, and \textbf{EpiQAL-C} tests conclusion reconstruction with the Discussion section withheld. Our contributions are as follows.
\begin{enumerate}[label=\textbullet, leftmargin=*, nosep]
    \item We frame epidemiological QA as a distinct problem requiring population-level reasoning over study evidence, and develop an expert-curated taxonomy ensuring broad subdomain coverage.
    \item We propose an automated construction framework integrating multi-LLM verification, difficulty control, and targeted human review.
    \item We release EpiQAL with three subsets and benchmark fifteen models spanning open-source and proprietary systems under a set-based evaluation protocol.
\end{enumerate}

\section{Related Work}
\label{sec:related}
 
\vskip 0.2em \noindent\textbf{Biomedical QA benchmarks.}
Existing biomedical QA benchmarks vary in format, evidence source, and domain scope. Exam-style benchmarks such as MedQA and MedMCQA use single-answer multiple-choice questions to test broad medical knowledge \citep{jin2021disease, pmlr-v174-pal22a}. BioASQ provides expert-curated questions with summaries and exact answers grounded in biomedical literature \citep{BioASQQA}, while PubMedQA links questions to abstracts but adopts a constrained yes/no/maybe label space that limits expressiveness \citep{jin-etal-2019-pubmedqa}. Epidemic-focused benchmarks such as COVID-QA, CoQUAD, and EPIC-QA ground questions in pandemic-related evidence but are typically disease-specific and use extractive formats \citep{moller-etal-2020-covid, CoQUADraza, epicqa}. In contrast, EpiQAL covers diverse epidemiological topics, supports set-based evaluation accommodating both single- and multi-answer questions, and includes a masked-input setting for conclusion reconstruction.
 
\vskip 0.2em \noindent\textbf{Automatic QA construction and quality control.}
Automatic QA construction has progressed from template-based generation \citep{du-etal-2017-learning} to adversarial filtering and model-in-the-loop collection that select harder, less biased instances \citep{Bartolo_2020, kiela2021dynabenchrethinkingbenchmarkingnlp, bras2020adversarialfiltersdatasetbiases, lee2025generatingplausibledistractorsmultiplechoice}. Multi-judge LLM verification mitigates single-model biases \citep{liu-etal-2023-g, ma2025judgingmindsperspectivesmean}, though judge models draw on the same parametric knowledge as the systems they assess, which has motivated verification signals read off an external corpus instead \citep{fan2026verifiablerewardsmathcode}. Set-based metrics support multi-answer evaluation \citep{yang-etal-2018-hotpotqa, 10.1609/aaai.v37i11.26529}, and difficulty screening further controls benchmark quality \citep{bai2024longbench2}.

\begin{figure*}[th!]
    \centering
    \includegraphics[width=0.9\textwidth, page=1, trim=5.7cm 2.4cm 6cm 2.6cm, clip]{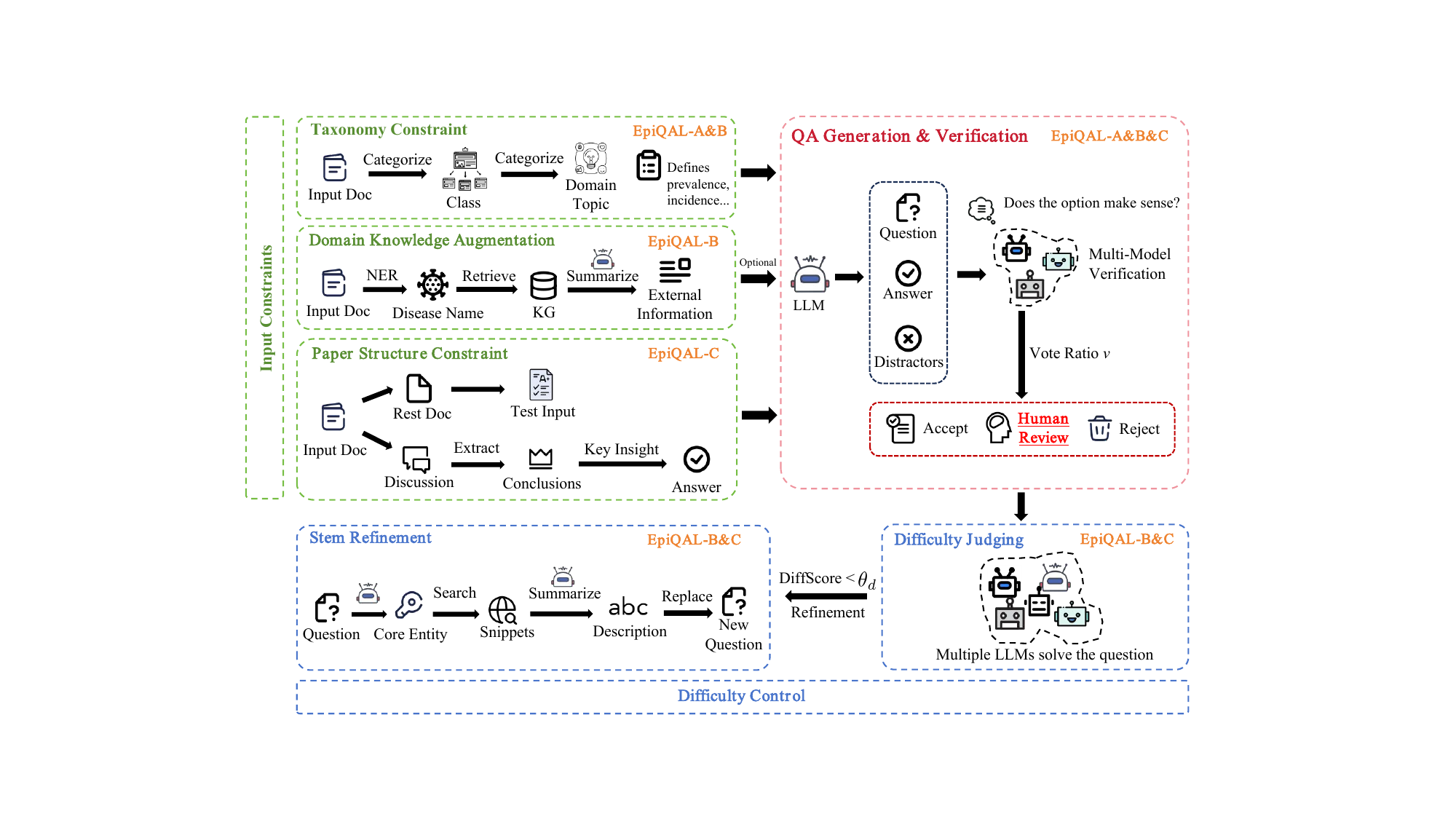}
    \caption{Overall framework for EpiQAL construction. The pipeline begins with subset-specific input processing (upper left), followed by QA generation and multi-model verification that routes uncertain cases to human review (upper right). For EpiQAL-B\&C, difficulty judging screens overly easy instances and triggers stem refinement when needed (lower). EpiQAL-A outputs directly after verification.}
    \label{fig:framework}
\end{figure*}

\begin{table*}[htbp]
\centering
\caption{Comparison of the three subsets in EpiQAL with representative examples.}
\small
\begin{tabular}{lccc}
\toprule
 & \textbf{EpiQAL-A} & \textbf{EpiQAL-B} & \textbf{EpiQAL-C} \\
\midrule
\textbf{Core Capability} & Fact recall & Multi-step inference & Conclusion reconstruction \\
\textbf{Knowledge Source} & Document & Document & Article body (Discussion masked) \\
\textbf{Taxonomy Guided} & Yes & Yes & No \\
\textbf{External Knowledge} & No & Generation only & No \\
\textbf{Test Input} & Full document & Full document & Document w/o Discussion \\
\textbf{Difficulty Control} & No & Yes & Yes \\
\bottomrule
\end{tabular}
\label{tab:subset_comparison}

\vspace{0.5em}

\small
\begin{tabular}{p{14.5cm}}
\toprule
\textbf{EpiQAL-A} \textbar{} \textbf{Q:} What was the primary outcome measure recommended by WHO for assessing anthelmintic drug efficacy in STH infections? \textbf{Opt:} (0) Number of stool samples taken (1) Cure rate (2) Log-transformed EPG (3) Egg reduction rate \textbf{Ans:} 3 \\
\midrule
\textbf{EpiQAL-B} \textbar{} \textbf{Q:} In experimental rabies infection in dogs, what most strongly distinguishes survivors from non-survivors in CNS immune response? \textbf{Opt:} (0) Viral antigens in CNS (1) High-titer serum VNA regardless of CSF (2) Elevated CSF WBC counts (3) VNA in CSF indicating immune effector entry into CNS (4) Earlier symptom onset \textbf{Ans:} 3 \\
\midrule
\textbf{EpiQAL-C} \textbar{} \textbf{Q:} What can be inferred about the overall impact of AVI-7288 treatment on Marburg virus infection in the macaque model? \textbf{Opt:} (0) Treatment may impair adaptive immunity (1) Marked improvement in survival and amelioration of disease manifestations (2) Treatment delayed clinical disease onset (3) Most effective within 24h despite similar outcomes (4) May not prevent viral reservoir formation \textbf{Ans:} 1 \\
\bottomrule
\end{tabular}
\label{tab:examples}
\end{table*}

\section{Method}

\subsection{Task Formulation}
\vskip 0.2em \noindent\textbf{Dataset generation.}
Given a source document $\mathcal{D}$, the goal is to produce a question $\mathcal{Q}$, a set of correct options $\mathcal{O}_c$, and a set of distractors $\mathcal{O}_d$. We formulate this as constrained generation where a model $\mathcal{M}_g$ operates under a constraint schema $\mathcal{G}$ that specifies topic scope, reasoning requirements, and option construction rules:
\begin{equation}
(\mathcal{Q}, \mathcal{O}_c, \mathcal{O}_d) = \mathcal{M}_g(\mathcal{D}, \mathcal{E}; \mathcal{G})
\end{equation}
Here $\mathcal{E}$ denotes optional external knowledge. For EpiQAL-B, $\mathcal{E}$ consists of epidemiological relations from knowledge graphs used only during construction, whose contribution is analyzed in Appendix~\ref{app:rag_analysis}; for EpiQAL-A and EpiQAL-C, $\mathcal{E}$ is empty. The constraint schema and its subset-specific instantiations are described in \S\ref{sec:subset_design}.
 
\vskip 0.2em \noindent\textbf{Benchmarking.}
The evaluation task is multiple choice QA. EpiQAL-A admits multiple correct options per question, while EpiQAL-B and EpiQAL-C are single-answer by construction. The evaluation prompt nonetheless states that a question may have one, several, or no correct options, so that answer cardinality is not disclosed to the model and cannot be used as a shortcut. Models therefore output a set of selected option indices in all cases. Let $\tilde{\mathcal{D}}$ denote the test-time input: $\tilde{\mathcal{D}} = \mathcal{D}$ for EpiQAL-A and EpiQAL-B, and $\tilde{\mathcal{D}} = \mathcal{D} \setminus \mathcal{D}_d$ for EpiQAL-C where the Discussion section $\mathcal{D}_d$ is masked. Given $\tilde{\mathcal{D}}$, question $\mathcal{Q}$, and candidates $\mathcal{O} = \mathcal{O}_c \cup \mathcal{O}_d$, a tested model $\mathcal{M}_t$ predicts an answer set $\mathcal{A} = \mathcal{M}_t(\tilde{\mathcal{D}}, \mathcal{Q}, \mathcal{O})$.
Evaluation uses set-based F1 and Exact Match (EM). Given predicted set $\mathcal{A}$ and reference set $\mathcal{O}_c$: $F_1 = 2|\mathcal{O}_c \cap \mathcal{A}| \,/\, (|\mathcal{O}_c| + |\mathcal{A}|)$ rewards partial overlap, while $\text{EM} = \mathbbm{1}[\mathcal{A} = \mathcal{O}_c]$ requires exact set recovery.

\subsection{Framework and Subset Design}
\label{sec:subset_design}
We design three subsets that target complementary capabilities along the epidemiological reasoning spectrum: text-grounded recall (EpiQAL-A), multi-step inference (EpiQAL-B), and conclusion reconstruction under masked inputs (EpiQAL-C). Figure~\ref{fig:framework} illustrates the shared construction pipeline, which proceeds from subset-specific input processing through constrained QA generation, multi-LLM verification with human review, and difficulty control. Table~\ref{tab:subset_comparison} summarizes the key differences across subsets.
 
\vskip 0.2em \noindent\textbf{EpiQAL-A: Text-grounded recall.}
EpiQAL-A contains retrieval-based questions whose correct options $\mathcal{O}_c$ are explicitly stated in the source document $\mathcal{D}$. Each correct option must be directly supported by verbatim spans. Distractors $\mathcal{O}_d$ are document-grounded confounders that match surface form but differ in role, population, or context.
 
\vskip 0.2em \noindent\textbf{EpiQAL-B: Multi-step inference.}
EpiQAL-B targets inference that links multiple cues in $\mathcal{D}$ with epidemiological knowledge. During construction, external knowledge $\mathcal{E}$ from knowledge graphs is optionally provided to supplement document evidence; at evaluation time $\mathcal{E}$ is withheld. Correct options $\mathcal{O}_c$ express derived implications rather than passage restatements. Distractors $\mathcal{O}_d$ contain reasoning-level flaws such as causal reversal (flipping the direction of an implied effect) or entity misattribution (applying a conclusion valid for one entity to a different one).
 
\vskip 0.2em \noindent\textbf{EpiQAL-C: Masked-input reasoning.}
EpiQAL-C evaluates reconstruction of author-stated conclusions when the Discussion section $\mathcal{D}_d$ is masked. Correct options $\mathcal{O}_c$ are salient conclusions extracted from $\mathcal{D}_d$, but must be supportable by evidence in $\tilde{\mathcal{D}}$. Distractors $\mathcal{O}_d$ are plausible under the paper narrative but unsupported, contradictory, or logically inverted.
 
Table~\ref{tab:examples} shows a representative question from each subset. Generation follows a constraint schema $\mathcal{G}$ specifying topic scope, reasoning type, and option construction rules for each subset; detailed distractor design and full prompts are in Appendices~\ref{app:distractors} and~\ref{app:prompt}.

\subsection{Input Constraints}
\label{sec:inputcon}
\vskip 0.2em \noindent\textbf{Epidemiology Taxonomy.}
Existing classification frameworks in epidemiology are organized along dimensions that do not suit our purpose. Study design taxonomies \citep{10.1093/ije/dys049} classify by research methodology, ICD-11 organizes diseases by pathology, and the CDC Field Epidemiology Manual structures knowledge by investigation procedure. None of these frameworks organize epidemiological knowledge by \emph{reasoning competency}, which is what we need for controlling question generation across distinct inference types. To ensure broad coverage across epidemiological subdomains, we develop a taxonomy in consultation with a faculty member in biostatistics and epidemiology through five rounds of iterative refinement, where each round involved reviewing example articles per topic and refining class/topic boundaries and their descriptions. The taxonomy is organized into six high-level classes covering complementary stages of epidemiological investigation. Each class contains multiple topics that provide finer-grained control over question intent. For EpiQAL-A and EpiQAL-B, we sample a topic and use its description to steer evidence selection, question phrasing, and option design. EpiQAL-C derives supervision from paper structure rather than taxonomy guidance. The complete taxonomy with all 25 topics and their descriptions is provided in Appendix~\ref{app:epi_tax}.

\vskip 0.2em \noindent\textbf{Domain Knowledge Retrieval.}
EpiQAL-B optionally incorporates external knowledge $\mathcal{E}$ from biomedical knowledge graphs during construction. When relevant knowledge graph triples are available, they provide additional context that may encourage broader inference-oriented questions. At evaluation time, $\mathcal{E}$ is withheld. Appendix~\ref{app:external_knowledge} details the construction procedure.

\subsection{Multi-model Verification}
\label{sec:modelverify}
Automatically generated QA instances may contain factual errors, label inconsistencies, or reasoning flaws. We address this through multi-model verification combined with targeted human review.

\vskip 0.2em \noindent\textbf{Checking model group.}
A group of LLMs independently verifies each generated option in $\mathcal{O}_c \cup \mathcal{O}_d$. Checkers assess two properties: whether the option is consistent with its assigned label given the cited evidence, and whether the implied reasoning is coherent. Checkers operate at the option level rather than re-solving the full question, which allows efficient verification at scale. In particular, checkers verify semantic equivalence between options, ensuring that trivially reordered or paraphrased variants of correct answers are not retained as distractors.

Checkers judge correctness solely against the test-time input $\tilde{\mathcal{D}}$, so that validity does not depend on construction-only information. For EpiQAL-C the checker additionally receives $\mathcal{D}_d$ in order to confirm that each extracted conclusion is traceable to the Discussion, but is instructed to decide derivability from $\tilde{\mathcal{D}}$ alone (Appendix \ref{app:prompt}).

We run each checker multiple times with stochastic decoding and aggregate decisions into a vote ratio $v \in [0,1]$ representing the fraction of keep votes. Two thresholds govern the decision process: options below the lower threshold are rejected automatically, options above the upper threshold are accepted, and options in between are flagged for human review (e.g., with 9 total votes, options receiving fewer than 5 keep votes are rejected, exactly 5 are flagged for human review, and 6 or more are accepted). This tiered approach balances automation with quality control.

\vskip 0.2em \noindent\textbf{Human Review.}
For flagged options, a human reviewer inspects the evidence to approve or discard. The reviewer may use LLMs for comprehension but all decisions are made by human judgment.

\subsection{Difficulty Control} 
\label{sec:difficulty}
For EpiQAL-B and EpiQAL-C, quality also depends on whether items demand nontrivial reasoning. We apply difficulty control only to these two subsets because EpiQAL-A targets text-grounded recall rather than reasoning depth. Difficulty control consists of two steps: difficulty judging to identify overly easy items, and stem refinement to reduce shortcut cues.

\vskip 0.2em \noindent\textbf{Difficulty judging.}
We estimate instance difficulty using a pool of models ranging from small to large. For each model, we compare the predicted answer set $\mathcal{A}$ with the reference set $\mathcal{O}_c$ using set-based F1 and Exact Match (Appendix~\ref{sec:appendix_option_set_metrics}), then combine them into a difficulty score:
\begin{equation}
  \label{eq:diffscore}
  \text{DiffScore} = 1 - \left( \alpha \cdot F_1 + (1-\alpha) \cdot \text{EM} \right) \nonumber
\end{equation}
where $\alpha \in [0,1]$ controls the trade-off between partial overlap and exact set recovery. We average DiffScore across the model pool. Items below a threshold are treated as easy and passed to stem refinement.

\vskip 0.2em \noindent\textbf{Stem refinement.}
Stem refinement is a rewriting step that replaces salient entities in the question stem $\mathcal{Q}$ with descriptive phrases. This reduces surface matching between $\mathcal{Q}$ and $\mathcal{O}_c$, requiring models to reason about the described concept rather than pattern match on entity names. The refinement procedure iteratively replaces salient entities with descriptive phrases retrieved from web sources, repeating until DiffScore exceeds the threshold or a maximum number of iterations is reached. No retrieved text is provided at evaluation time. We cap iterations at $T_r = 3$ with early stopping, as later iterations can over-expand entity descriptions and partially offset the intended difficulty increase. The detailed procedure, an example, and performance analysis are in Appendices~\ref{app:stem_refine_details} and~\ref{app:stem_refine_analysis}. Refinement touches a minority of items, since 317 of 478 EpiQAL-B instances keep their original stem (Appendix~\ref{app:difficulty_analysis}). We therefore treat it as one quality-control step for reducing lexical shortcuts on the easiest items, not as a central mechanism of the benchmark.


\section{Experiment}
\subsection{Generation Settings}
\vskip 0.2em \noindent\textbf{Generation and verification.}
We use Qwen3-30B-A3B-Instruct-2507 as the generation model. For EpiQAL-B, disease entities are linked to biomedical knowledge graphs (Appendix~\ref{app:external_knowledge}) to optionally supplement generation; post-hoc analysis found this knowledge contributed minimally (Appendix~\ref{app:rag_analysis}). Generated options are verified by three models from different families (GPT-5-mini, DeepSeek-V3.2-Thinking, GLM-4.5-Air) with vote-ratio aggregation. Difficulty control uses a four-model pool (GPT-5-mini, DeepSeek-V3.2-Thinking, Qwen3-32B, Phi-4-mini-instruct) with $\alpha = 0.3$ and threshold $\theta_d = 0.2$. Implementation details, DiffScore distribution, and threshold selection rationale are in Appendices~\ref{app:exp_details} and~\ref{app:difficulty_analysis}.

\vskip 0.2em \noindent\textbf{Corpus.}
We build a corpus of approximately 10,600 research articles from the Journal Archive of PLOS Neglected Tropical Diseases \cite{PLOS_NTD_Journal}, which covers over 20 diseases across viral, parasitic, bacterial, and fungal pathogens with consistent section formatting under a CC BY 4.0 license. We use a randomly sampled subset of 500 articles for the main experiments.

Table~\ref{tab:stats} summarizes dataset statistics.
After multi-model verification and human review of ambiguous options, instances where all correct options were rejected are discarded, yielding 475, 478, and 479 instances for EpiQAL-A, B, and C respectively.
The verification process is detailed in \S\ref{sec:quality}.

\subsection{Benchmark Quality Evaluation}
\label{sec:quality}
We evaluate benchmark quality through automated verification statistics and human evaluation of the finalized dataset, with a concrete verification example in Appendix~\ref{app:checker_example}.

\subsubsection{Verification Pipeline Statistics}
Each checker runs 3 times with temperature 1.0, yielding 9 total votes per option. For flagged options, a human reviewer inspects the evidence and votes to approve or discard; all decisions are made by human judgment. Table~\ref{tab:stats} reports verification outcomes across the three subsets.

\begin{table}[h]
\caption{Dataset statistics and verification outcomes.}
\centering
\small
\begin{tabular}{lccc}
\toprule
\textbf{EpiQAL} & \textbf{A} & \textbf{B} & \textbf{C} \\
\midrule
Samples & 475 & 478 & 479 \\
Avg. \#Options & 3.21 & 4.90 & 4.73 \\
Avg. \#Correct & 1.19 & 1.00 & 1.00 \\
\midrule
\multicolumn{4}{l}{\textit{Option-Level}} \\
\quad Accept & 86.3\% & 94.1\% & 89.7\% \\
\quad Reject & 9.8\% & 4.4\% & 7.5\% \\
\quad Human Review & 3.9\% & 1.5\% & 2.8\% \\
\midrule
\multicolumn{4}{l}{\textit{Instance-Level}} \\
\quad All Accepted & 65.0\% & 78.0\% & 67.6\% \\
\quad Partial Reject & 20.4\% & 13.4\% & 18.6\% \\
\quad Needs Review & 11.0\% & 6.2\% & 11.6\% \\
\quad Discarded & 3.6\% & 2.4\% & 2.2\% \\
\bottomrule
\end{tabular}
\label{tab:stats}
\end{table}

The pipeline automatically resolves over 96\% of options without human intervention, with only 2.2--3.6\% of instances fully discarded. Appendix~\ref{app:checker_example} illustrates the checker detecting semantic equivalence between superficially distinct options.

\subsubsection{Human Evaluation}
\label{sec:human_eval}
Three computer science PhD students with biomedical training independently score a stratified sample of 120 questions (40 per subset) after benchmark construction is complete. Annotators may use LLMs to aid comprehension but all scoring is performed by human judgment.
For EpiQAL-B and EpiQAL-C, we stratify by difficulty: 12 Easy (DiffScore = 0), 14 Medium (0 < DiffScore $\leq$ 0.4), and 14 Hard (DiffScore > 0.4).
For EpiQAL-A, we use random sampling since difficulty screening is not applied.

Annotators rate each question on subset-specific dimensions using a 1--3 scale (1 = poor, 2 = acceptable, 3 = good). As shown in Table~\ref{tab:human_eval}, all dimensions score above 2.0 across subsets, with Answer Correctness ($\geq$2.81) and Evidence Sufficiency ($\geq$2.75) scoring highest. Exact Agreement on the 3-point scale ranges from 47.5\% to 68.8\%; however, Binary Agreement (acceptable vs.\ poor) exceeds 91\% on all subsets and Severe Disagreement (items where both 1 and 3 appear) stays below 8\%, indicating that most variation reflects fine-grained distinctions between acceptable and good rather than genuine quality disputes.

A faculty member in biostatistics and epidemiology scored 30 questions (10 per subset) using the same dimensions and scale, blind to the annotators' scores. Binary Agreement against annotator majority votes reaches 97.5\% (A), 96.0\% (B), and 98.3\% (C), with Severe Disagreement at 2.5\%, 2.0\%, and 0.0\%, indicating that annotator judgments track the expert on whether an item is acceptable. This sample is small and does not establish that every automatically verified label has been expert-validated. Detailed protocol and per-dimension analysis are in Appendix~\ref{app:human_eval}.

\begin{table}[h]
\caption{Human evaluation scores (mean across annotators, 1--3 scale) and inter-annotator agreement.}
\centering
\small
\begin{tabular}{lccc}
\toprule
\textbf{Dimension} & \textbf{A} & \textbf{B} & \textbf{C} \\
\midrule
Answer Correctness & 2.86 & 2.81 & 2.92 \\
Distractor Quality & 2.52 & 2.53 & 2.56 \\
Question Clarity & 2.77 & 2.65 & 2.51 \\
Evidence Sufficiency & 2.93 & 2.75 & 2.77 \\
Reasoning Depth & -- & 2.45 & 2.45 \\
Answerability & -- & -- & 2.78 \\
\midrule
Exact Agreement (\%) & 68.8 & 47.5 & 55.4 \\
Binary Agreement (\%) & 95.0 & 91.5 & 91.7 \\
Severe Disagreement (\%) & 3.1 & 7.5 & 5.0 \\
\bottomrule
\end{tabular}
\label{tab:human_eval}
\end{table}

\subsection{Model Evaluation}
\label{sec:results}

\subsubsection{Setup}

We evaluate 15 models from seven families: OpenAI \citep{singh2025openaigpt5card, openai2024gpt4technicalreport}, DeepSeek \citep{deepseekai2025deepseekv32pushingfrontieropen, deepseekai2026deepseekv4}, Zhipu AI \citep{5team2025glm45agenticreasoningcoding}, Microsoft \citep{microsoft2025phi4minitechnicalreportcompact}, Meta \citep{grattafiori2024llama3herdmodels}, Mistral AI \citep{jiang2023mistral7b}, and Alibaba-Qwen \citep{yang2025qwen3technicalreport}, ranging from 3B to over 100B parameters (full list in Table~\ref{tab:bench}). After our initial experiments, the DeepSeek-V3.2-Thinking API was discontinued and replaced by V4-Flash-Thinking; we include both in the main evaluation, and appendix experiments use whichever version was available at the time. All models receive $\tilde{\mathcal{D}}$, $\mathcal{Q}$, and $\mathcal{O}$ in a single prompt. Models output a JSON object with selected option indices; we use vLLM structured output for local models and regex post-processing for API models. Temperature is 0.3 except for reasoning models (1.0). We choose 0.3 to balance reproducibility with output diversity, avoiding degenerate repetitions sometimes observed at T=0. An ablation across 12 models confirms that switching to T=0 changes mean per-model EM by less than 1\%, with Spearman rank correlation exceeding 0.99 on all subsets (Appendix~\ref{app:temp_sensitivity}). A multi-run analysis further confirms that results are stable across three runs, with mean standard deviation under 1\% (Appendix~\ref{app:multi_run}). We report set-based F1 and Exact Match under zero-shot no-CoT, zero-shot CoT, and one-shot settings (Appendix~\ref{app:prompt}).

\subsubsection{Results}
Table~\ref{tab:bench} reports Exact Match and F1 on all three subsets under zero-shot no-CoT and CoT settings. Exact Match is a pass rate over roughly 478 items per subset, so 95\% Wilson intervals have a half-width of at most about 0.045. For instance, DeepSeek-V4-Flash-Thinking on EpiQAL-B is 0.703 with an interval of 0.661 to 0.742.

\begin{table*}[htbp]
    \centering 
    \small
    \caption{Exact Match|F1 Score for each model across subsets, with and without Chain-of-Thought prompting. \textbf{Bold} = best, \underline{underline} = second best per column. Ties share the same mark.}
    \label{tab:bench}
    \begin{tabular}{lcccccc}
        \toprule
         & \multicolumn{2}{c}{\textbf{EpiQAL-A}} & \multicolumn{2}{c}{\textbf{EpiQAL-B}} & \multicolumn{2}{c}{\textbf{EpiQAL-C}} \\
        \cmidrule(lr){2-3} \cmidrule(lr){4-5} \cmidrule(lr){6-7}
        \textbf{Model} & w/o CoT & CoT & w/o CoT & CoT & w/o CoT & CoT \\
        \midrule
        \textit{OpenAI} &&&&&&\\
        \hspace*{1em}GPT-5-mini                     & 0.905|0.966 & 0.924|0.966 & 0.533|0.788 & 0.634|0.827 & 0.599|0.789 & 0.555|0.762 \\
        \hspace*{1em}GPT-4o-mini                    & 0.766|0.904 & 0.796|0.913 & 0.222|0.650 & 0.531|0.760 & 0.213|0.655 & 0.236|0.661 \\
        \hspace*{1em}GPT-4.1-nano                   & 0.768|0.855 & 0.792|0.868 & 0.678|0.809 & 0.642|0.789 & 0.559|0.791 & 0.553|0.780 \\
        \midrule
        \textit{DeepSeek} &&&&&&\\
        \hspace*{1em}DeepSeek-V3.2-Thinking          & \textbf{0.928}|\underline{0.970} & \underline{0.928}|\underline{0.969} & \textbf{0.818}|\textbf{0.896} & \textbf{0.868}|\textbf{0.909} & \underline{0.720}|0.804 & \underline{0.716}|0.822 \\
        \hspace*{1em}DeepSeek-V4-Flash-Thinking      & \underline{0.926}|\textbf{0.971} & \textbf{0.935}|\textbf{0.970} & 0.703|0.841 & 0.795|0.862 & 0.666|0.785 & 0.697|0.794 \\
        \midrule
        \textit{Microsoft} &&&&&&\\
        \hspace*{1em}Phi-4-mini-instruct             & 0.583|0.811 & 0.657|0.819 & 0.249|0.678 & 0.406|0.736 & 0.426|0.755 & 0.388|0.744 \\
        \midrule
        \textit{Meta-Llama} &&&&&&\\
        \hspace*{1em}Llama-3.2-3B-Instruct           & 0.366|0.553 & 0.339|0.473 & 0.157|0.511 & 0.100|0.259 & 0.088|0.375 & 0.113|0.366 \\
        \hspace*{1em}Llama-3.1-8B-Instruct           & 0.798|0.911 & 0.834|0.920 & 0.318|0.670 & 0.665|0.799 & 0.190|0.592 & 0.384|0.708 \\
        \hspace*{1em}Llama-3.3-70B-Instruct          & 0.779|0.884 & 0.789|0.890 & 0.651|0.825 & 0.732|0.855 & 0.580|0.805 & 0.626|\underline{0.827} \\
        \midrule
        \textit{Mistral AI} &&&&&&\\
        \hspace*{1em}Mistral-7B-Instruct-v0.3        & 0.722|0.809 & 0.709|0.799 & \underline{0.789}|0.808 & \underline{0.812}|0.814 & \textbf{0.808}|\underline{0.822} & \textbf{0.812}|0.816 \\
        \hspace*{1em}Mistral-Large-Instruct-2411     & 0.901|0.956 & 0.916|0.960 & 0.644|0.844 & 0.688|0.853 & 0.685|\textbf{0.856} & 0.699|\textbf{0.856} \\
        \midrule
        \textit{Qwen} &&&&&&\\
        \hspace*{1em}Qwen3-8B                        & 0.811|0.921 & 0.840|0.929 & 0.508|0.763 & 0.619|0.816 & 0.484|0.752 & 0.514|0.768 \\
        \hspace*{1em}Qwen3-30B-A3B-Instruct-2507     & 0.882|0.956 & 0.899|0.964 & 0.573|0.807 & 0.747|\underline{0.873} & 0.585|0.807 & 0.622|0.826 \\
        \hspace*{1em}Qwen3-32B                       & 0.872|0.949 & 0.857|0.943 & 0.743|\underline{0.864} & 0.736|0.853 & 0.547|0.783 & 0.557|0.779 \\
        \midrule
        \textit{Zhipu AI} &&&&&&\\
        \hspace*{1em}GLM-4.5-Air                     & 0.874|0.947 & 0.884|0.953 & 0.657|0.836 & 0.655|0.835 & 0.580|0.782 & 0.572|0.733 \\
        \bottomrule
    \end{tabular}
\end{table*}

\subsection{Distractor Error Analysis}
\label{sec:error}
We classify each distractor by its deception mechanism using GPT-5-mini and cross-reference with model misselections under zero-shot no-CoT. Manual verification of a random sample of 15 questions per subset confirmed that the automatically assigned distractor categories were consistent with human judgment in inspected cases. Each subset uses four categories reflecting its distractor design: EpiQAL-A distinguishes wrong entity/role, wrong context, wrong metric, and semantic near-miss; EpiQAL-B distinguishes methodological mismatch, causal/logical error, variable confusion, and assumption violation; EpiQAL-C distinguishes external dependency, speculation/limitation, background only, and causal reversal. Full categories and distractor distributions are provided in Appendix~\ref{app:distractor_analysis}.

\begin{table}[h]
\caption{Distractor deception rate by category.}
\centering
\small
\renewcommand{\arraystretch}{0.85}
\begin{tabular}{llc}
\toprule
\textbf{EpiQAL} & \textbf{Category} & \textbf{Rate} \\
\midrule
\multirow{4}{*}{\textbf{A}} & Semantic near-miss & 16.8\% \\
& Wrong metric & 9.4\% \\
& Wrong context & 8.1\% \\
& Wrong entity/role & 8.0\% \\
\midrule
\multirow{4}{*}{\textbf{B}} & Variable confusion & 19.2\% \\
& Causal/logical error & 17.7\% \\
& Methodological mismatch & 17.0\% \\
& Assumption violation & 15.9\% \\
\midrule
\multirow{4}{*}{\textbf{C}} & Speculation/limitation & 21.0\% \\
& External dependency & 13.8\% \\
& Causal reversal & 12.6\% \\
& Background only & 39.3\%$^\dagger$ \\
\bottomrule
\end{tabular}

\vspace{0.3em}
\raggedright\footnotesize{\hspace{1.5em}$\dagger$Based on only 8 distractors; rate is unreliable.}
\label{tab:deception_rate}
\end{table}

We define the \textit{deception rate} of a distractor category as the fraction of times it is misselected, computed over the 14 models available at the time of this analysis (Appendix~\ref{app:distractor_analysis}). Table~\ref{tab:deception_rate} reports deception rates across all three subsets. On EpiQAL-A, Semantic near-miss distractors are roughly twice as deceptive as other categories. On EpiQAL-B, rates are uniformly high (15.9--19.2\%), and on EpiQAL-C, Speculation/limitation leads at 21.0\%. Background only is higher at 39.3\% but is based on only 8 distractors and is therefore excluded from this comparison. Per-model rates are in Appendix~\ref{app:permodeldecrate}.

\subsection{Discussion}
\label{sec:discussion}

\vskip 0.2em \noindent\textbf{Epidemiological reasoning remains challenging.}
DeepSeek-V3.2-Thinking, the strongest model overall, achieves 0.928 EM on EpiQAL-A with CoT, essentially tied with DeepSeek-V4-Flash-Thinking at 0.935, and leads EpiQAL-B at 0.868, while Mistral-7B leads EpiQAL-C at 0.812, indicating that epidemiological reasoning over full research articles poses a distinct challenge.

\vskip 0.2em \noindent\textbf{Document evidence matters, though grounding is weaker on EpiQAL-B.}
Input ablations (Appendix~\ref{app:input_ablation}) feed models progressively less context. On EpiQAL-A and EpiQAL-C, replacing the correct article with a random one from the same subset drops Exact Match to near zero, below the question-only baseline, indicating that models condition on the article rather than on the options alone. Option-only accuracy does not exceed the empirical single-pick reference on any subset (0.31 on A, 0.21 on B, 0.22 on C), so the options are not self-revealing in isolation, though this does not rule out a shortcut that uses the question and options together. EpiQAL-B is mixed: DeepSeek-V4-Flash-Thinking and GPT-4.1-nano gain 0.196 and 0.240 EM from the correct passage over question-only, while GPT-5-mini and GPT-4o-mini show little measurable benefit. Residual question-only performance on B and C is comparable before and after stem refinement (B: 0.353 vs.\ 0.356; C: 0.253 vs.\ 0.246), so it is not a refinement artifact, but the present controls cannot separate question-option informativeness from parametric knowledge. We therefore treat evidence grounding on EpiQAL-B as partial and revisit it in the Limitations. An OOD test set from 96 post-release articles yields Spearman rank correlation at least 0.94 on all subsets (Appendix~\ref{app:ood_analysis}), indicating that rankings are not driven by memorization of training-set passages. A cross-source evaluation on 82 articles from the International Journal of Epidemiology, which covers broader epidemiological topics including chronic and environmental epidemiology, reproduces the main qualitative patterns. Mistral-7B leads on EpiQAL-C, Llama-3.2-3B collapses on the reasoning subsets, and scale does not predict success (Appendix~\ref{app:cross_source}). Because this uses the same generator and checkers with no separate expert pass, we read it as evidence that the construction pipeline ports beyond neglected tropical diseases rather than as coverage of epidemiology as a whole.

\vskip 0.2em \noindent\textbf{EpiQAL-B is empirically the most challenging subset under our evaluation setup, and model rankings suggest partially distinct behaviors across subsets.}
Without CoT, only four of fifteen models exceed 0.70 EM on EpiQAL-B, and deception rates are uniformly high across all distractor categories (15.9--19.2\%). By comparison, twelve exceed 0.76 on EpiQAL-A and ten exceed 0.50 on EpiQAL-C. Spearman correlation across the fifteen models (zero-shot no-CoT) yields $\rho = 0.482$ for A vs.\ B and $\rho = 0.615$ for A vs.\ C, while B vs.\ C reaches $\rho = 0.760$, suggesting that the subsets capture partially distinct model behaviors.

\vskip 0.2em \noindent\textbf{Model rankings are subset-dependent, and scale alone does not predict success.}
DeepSeek-V3.2-Thinking ranks first on A and B but falls to second on C (0.720), behind Mistral-7B (0.808). Mistral-7B also outperforms Mistral-Large on both B (0.789 vs.\ 0.644) and C (0.808 vs.\ 0.685) despite having a fraction of the parameters. Llama-3.2-3B collapses on reasoning-intensive subsets (0.157 on B, 0.088 on C), suggesting that a capability threshold exists below which models cannot perform epidemiological reasoning; above it, instruction tuning quality and architecture may matter more than raw scale. Two models (Llama-3.3-70B and GLM-4.5-Air) use four-bit inference for memory reasons (Appendix~\ref{app:exp_details}), so their absolute scores may be slightly depressed relative to full precision. The scale claim does not rest on them: Mistral-Large and Mistral-7B are both run at default precision and differ by more than an order of magnitude in parameters. The two DeepSeek versions further illustrate this distinction: V4-Flash-Thinking nearly matches V3.2-Thinking on EpiQAL-A (0.926 vs.\ 0.928) but drops substantially on B (0.703 vs.\ 0.818) and C (0.666 vs.\ 0.720), consistent with the finding that a lighter model variant retains retrieval performance while disproportionately losing reasoning capability (Appendix~\ref{app:model_version}). The generation model Qwen3-30B-A3B ranks ninth on B (0.573), indicating it does not systematically outperform others on the benchmark it produced.

\vskip 0.2em \noindent\textbf{Distractor analysis reveals fine-grained failure modes.}
The F1--EM gap exposes systematic over-selection: on EpiQAL-B without CoT, Phi-4-mini ($\Delta = 0.429$) and GPT-4o-mini ($\Delta = 0.428$) frequently select distractors alongside correct options, whereas Mistral-7B shows near-zero gaps ($\Delta = 0.019$ on B, 0.014 on C), achieving strong EM through answer precision rather than broad coverage. In epidemiological applications where false positives carry costs such as unnecessary interventions, selective models may outperform those that maximize coverage. Per-model deception rates confirm that over-selection is driven by vulnerability to specific distractor categories rather than random guessing (Appendix~\ref{app:permodeldecrate}).

\vskip 0.2em \noindent\textbf{Prompting strategies.}
CoT yields the largest gains on EpiQAL-B, with Llama-3.1-8B improving from 0.318 to 0.665 EM ($+$0.347) and GPT-4o-mini from 0.222 to 0.531 ($+$0.309). On EpiQAL-A, gains are marginal. On EpiQAL-C, CoT is mixed: Llama-3.1-8B benefits (0.190 $\to$ 0.384) but GPT-5-mini degrades (0.599 $\to$ 0.555), suggesting that explicit reasoning chains can amplify errors under incomplete evidence. One-shot prompting uniformly improves EpiQAL-B, with the largest gains exceeding $+$0.3 (Appendix~\ref{app:1shot}).

\section{Conclusion}
We introduced EpiQAL, a benchmark that operationalizes evidence-grounded epidemiological reasoning over research articles. Our construction framework combines an expert-curated taxonomy, subset-specific constraints for evidence grounding, multi-model verification, and difficulty screening. This yields three complementary subsets targeting text-grounded recall, multi-step inference, and conclusion reconstruction.
Experiments across fifteen models spanning open-source and proprietary systems reveal that current LLMs struggle with epidemiological reasoning, with multi-step inference posing the greatest challenge. Model rankings shift across subsets, and scale alone does not predict success. Chain-of-Thought prompting benefits multi-step inference but yields mixed results elsewhere. Distractor error analysis identifies fine-grained failure modes, with reasoning-level distractors challenging models uniformly on multi-step inference while semantic near-misses dominate retrieval errors. A cross-source evaluation on the International Journal of Epidemiology indicates the pipeline generalizes beyond neglected tropical diseases, and we release EpiQAL to facilitate future work on evidence-grounded public health reasoning.
 
\section*{Limitations}
This work has several limitations. First, our controls do not separate epidemiological reasoning from general scientific reading. The input ablations (Appendix~\ref{app:input_ablation}) show that the specific article matters and that options alone are not self-revealing, which is weaker than showing that success requires epidemiology-specific knowledge. Establishing that target would require controls we did not run: a matched non-epidemiology benchmark built by the same pipeline, a comparison between epidemiologist and generalist human performance, and per-item annotation of which epidemiological principle each question depends on. Relatedly, two of the four API models gain little measurable accuracy from the correct passage on EpiQAL-B, so our evidence-grounding claims hold most clearly for EpiQAL-A and EpiQAL-C and only partially for EpiQAL-B. Cross-subset comparisons are also confounded by format. EpiQAL-A averages 3.21 options and admits multiple correct answers, while EpiQAL-B and EpiQAL-C average 4.90 and 4.73 options and are single-answer, so lower Exact Match on B and C may partly reflect option count and answer cardinality rather than reasoning difficulty alone. Second, our source corpus is drawn solely from PLOS Neglected Tropical Diseases and all content is in English, which may underrepresent other epidemiological domains and multilingual settings. The cross-source check on the International Journal of Epidemiology uses the same generator and checkers, so it tests pipeline portability rather than coverage of epidemiology as a whole. Third, we generate approximately 500 instances per subset using a single 30B-parameter model. A preliminary check with a stronger cross-family generator on a 50-article sample per subset largely preserves the per-model rankings (Appendix \ref{app:generator_ablation}), but it uses one alternative generator on a small sample and does not include a matched comparison of distractor deception rates, so a controlled generator ablation at full scale remains future work. Fourth, despite multi-model verification and targeted human review, residual errors may remain. Additionally, stem refinement can over-expand at later iterations, partially offsetting the intended difficulty increase (Appendix \ref{app:stem_refine_analysis}). Our human annotators are computer science PhD students with biomedical training rather than practicing epidemiologists, and our human evaluation covers 120 questions (8.4\% of the benchmark) scored by three annotators, with 30 questions scored by a faculty member in biostatistics and epidemiology. The rest of the benchmark rests on automated verification. Some evaluated models also participate in construction as checkers or difficulty judges. This overlap is unlikely to introduce answer leakage, since checkers only perform option-level validity filtering, while difficulty judges only identify overly easy items before a separate stem-refinement step. More broadly, generation, verification, difficulty judging, and the post-hoc distractor classification are all performed by LLMs. We use models from different families, but these systems can share instruction-tuning preferences and blind spots, so item style and difficulty remain model-relative to some degree (Appendix~\ref{app:role_overlap}). Fifth, because all source articles are publicly available, models trained after our corpus collection date may have encountered these passages during pre-training. We mitigate this concern through an out-of-distribution evaluation on post-release articles (Appendix~\ref{app:ood_analysis}), which preserves model rankings and argues against ranking effects driven by memorization. However, we cannot fully rule out indirect contamination through derivative training data. Finally, EpiQAL remains a proxy for real-world public health analysis, which often requires integrating multiple documents and incorporating temporal and geographic context beyond single-article reasoning.

\section*{Ethical Considerations}
The main benchmark is built from PLOS Neglected Tropical Diseases, which is open access under CC BY 4.0, and we release EpiQAL under terms consistent with that license. The cross-source analysis in Appendix~\ref{app:cross_source} uses articles from the International Journal of Epidemiology, a hybrid journal. These instances are not part of the released benchmark and are reported only as an experimental check. The benchmark contains no personally identifying information, since it is built from published research articles rather than clinical records.

Human annotation was performed by researchers involved in this project and by a faculty collaborator, all of whom participated with full knowledge of how their judgments would be used. No crowdworkers were recruited and no compensation issues arise.

EpiQAL is intended for evaluating language models on epidemiological reasoning over research literature. It is not a clinical or public health decision-making tool, and performance on this benchmark should not be read as evidence that a model is suitable for guiding real epidemiological practice. We note this because strong benchmark scores can be misread as domain competence, particularly in a health-adjacent setting.

\section*{Acknowledgments}
This research is partially supported by the U.S. National Science Foundation under Award Number 2504088, and the National Institute of Allergy and Infectious Diseases of the NIH under Award Numbers R01AI189829 and R01AI197111. The content is the sole responsibility of the authors and does not necessarily represent the views of the NIH. This project is also partially supported by the NAIRR Pilot Program.

We are grateful for the computing resources provided by the iTiger GPU cluster \citep{sharif2026cultivatingmultidisciplinaryaiworkforce}, supported by the NSF MRI program.

We thank the anonymous reviewers and the ARR action editors and chairs for their thoughtful feedback, which substantially improved this work.

\bibliography{custom}

\appendix
\section{Additional Related Work}
\label{app:relatedwork}

 \vskip 0.2em \noindent\textbf{Machine reading comprehension.} Early work on machine reading comprehension cast question answering as span selection within controlled contexts, enabling precise evaluation of extractive models \citep{rajpurkar-etal-2016-squad, joshi-etal-2017-triviaqa}. With the rise of instruction-tuned large language models, generation-based QA has become competitive, yet multiple choice formats remain attractive because they encourage targeted reasoning while preserving objective scoring \citep{nie-etal-2020-adversarial, hendrycks2021measuringmassivemultitasklanguage, zhang2024tacco}. Scientific articles often restate conclusions with considerable lexical overlap, meaning that purely extractive setups can overestimate genuine inference \cite{sugawara2018makes, wang2025biomedjimpact}. This observation motivates evaluation formats that probe reasoning beyond surface matching.

\vskip 0.2em \noindent\textbf{Additional biomedical QA resources.} Beyond the benchmarks discussed in the main text, several resources address specific clinical needs. emrQA constructs QA pairs from electronic medical records using expert templates \citep{pampari2018emrqalargecorpusquestion}. MedQuAD compiles question-answer pairs from trusted medical websites organized by topic \citep{BenAbacha-BMC-2019}. These datasets primarily target patient-level clinical reasoning rather than population-level epidemiological inference.

 \vskip 0.2em \noindent\textbf{Retrieval augmentation and knowledge resources.} Retrieval-augmented generation grounds model outputs in retrieved passages and is often used to mitigate hallucination \citep{10.5555/3495724.3496517, izacard-grave-2021-leveraging, bhasuran2025preliminary, wu2025utilizing}. Structured resources such as Hetionet and iBKH encode biomedical entities and relations that can support downstream reasoning \citep{10.7554/eLife.26726, Su2021.03.12.21253461}. For epidemiology-oriented knowledge, eKG-DONs compiles outbreak reports from official sources \citep{JRC139580}. Systematic evaluations of LLMs on knowledge base question answering have found that they still trail specialized models on structured reasoning tasks \citep{tan2023chatgptreplacetraditionalkbqa}. Recent work studies instruction-aware retrieval across heterogeneous sources \citep{min-etal-2025-unihgkr} and integration of knowledge graphs with multi-agent reasoning \citep{xie2025kerapknowledgeenhancedreasoningapproach, xie2025hypkg, xu-etal-2025-simrag}. In EpiQAL-B construction, we operationalize structured relations by summarizing knowledge graph triples into natural language signals used only during generation; these signals are withheld at evaluation time.

\section{Method Details}
\label{app:method_details}

\subsection{Evaluation Metrics}
\label{sec:appendix_option_set_metrics}

Let $set_{\text{model}}$ denote the set of options predicted by a model and $set_{\text{ref}}$ denote the reference option set. We compute
\begin{equation}
  \label{eq:f1score}
  F_1 = \frac{2 \cdot \left| set_{\text{reference}} \cap set_{\text{model}} \right|}{\left| set_{\text{reference}} \right| + \left| set_{\text{model}} \right|}
\end{equation}
\begin{equation}
  \label{eq:emscore}
  \text{ExactMatch} = 
    \begin{cases}
      1, & \text{if } set_{\text{model}}=set_{\text{reference}} \\
      0, & \text{otherwise}
    \end{cases}
\end{equation}

\subsection{Epidemiology Taxonomy}
\label{app:epi_tax}
This appendix provides the complete taxonomy introduced in Section~\ref{sec:inputcon}. Each of the six classes contains multiple topics, and each topic includes an expert-curated description specifying its semantic scope. These descriptions serve as explicit constraints during question generation for EpiQAL-A and EpiQAL-B, steering the generation model toward the intended epidemiological competency. The released instances carry class and topic labels, which support topic-level analysis in future work.

Table~\ref{tab:taxonomy_classes} lists the six classes with their descriptions. Tables~\ref{tab:taxonomy_topics_12} through \ref{tab:taxonomy_topics_56} provide all 25 topics organized by class.

\newcolumntype{L}[1]{>{\raggedright\arraybackslash}p{#1}}

\begin{table*}[t]
\centering
\caption{Epidemiology taxonomy classes.}
\footnotesize
\setlength{\tabcolsep}{5pt}
\renewcommand{\arraystretch}{1.12}
\begin{tabularx}{\textwidth}{L{0.05\textwidth} L{0.26\textwidth} X}
\toprule
\textbf{Cls} & \textbf{Class} & \textbf{Description} \\
\midrule
1 & Surveillance and Descriptive Epidemiology &
Describes population occurrence from routine data, including rates, time place person patterns, aberration signals, and basic system performance, without causal analysis or forecasting. \\
2 & Outbreak Investigation and Field Response &
Handles outbreak specific confirmation, field case definitions, line lists, attack rates and curves, chain and source hypotheses, and immediate control with situation reports. \\
3 & Determinants and Exposures &
Explains how exposure arises across settings, covering behavioral, environmental, occupational, and social determinants, delineates canonical transmission routes and contact structures, interprets exposure response with attention to measurement methods, units, detection limits, and thresholds, and situates risks within One Health interfaces involving reservoirs and vectors. \\
4 & Susceptibility and Immunity &
Describes who is susceptible and why, links serologic measures to correlates of protection, evaluates effectiveness after vaccination or prior infection and its waning with reinfection, hybrid immunity, and variant escape, including the effects of vaccine dose number and intervals, and assesses severity risk using clinical and contextual prognostic factors. \\
5 & Modeling, Methods, and Evaluation &
Provides analytical methods for transmission modeling and inference, real time debiasing of surveillance data, study design and causal effects, measurement and bias handling, and program performance and burden evaluation. \\
6 & Projections and Forecasts &
Produces forward looking forecasts and scenarios, evaluates and combines models, and supports decision making, it does not reconstruct recent under reported data. \\
\bottomrule
\end{tabularx}
\label{tab:taxonomy_classes}
\end{table*}

\begin{table*}[t]
\centering
\caption{Epidemiology taxonomy topics (Classes 1 and 2).}
\footnotesize
\setlength{\tabcolsep}{5pt}
\renewcommand{\arraystretch}{1.12}
\begin{tabularx}{\textwidth}{L{0.04\textwidth} L{0.22\textwidth} L{0.04\textwidth} L{0.22\textwidth} X}
\toprule
\textbf{Cls} & \textbf{Class} & \textbf{Top} & \textbf{Topic} & \textbf{Description} \\
\midrule

\multirow[t]{4}{*}{1} & \multirow[t]{4}{=}{Surveillance and Descriptive Epidemiology}
& 1 & Frequency measures and standardization
& Defines prevalence, incidence, person time, and applies standardization to make rates comparable. \\
& & 2 & Time Place Person patterns, seasonality and clustering
& Describes temporal trends, spatial distribution, and demographic profiles using routine population surveillance. \\
& & 3 & Aberration and outbreak detection
& Builds statistical baselines and thresholds to flag unusual increases in counts, rates, or positivity, focuses on signal detection rather than source attribution. \\
& & 4 & System performance, deduplication and record linkage
& Assesses sensitivity, timeliness, and completeness, manages deduplication and linkage across multiple data sources. \\
\addlinespace

\multirow[t]{4}{*}{2} & \multirow[t]{4}{=}{Outbreak Investigation and Field Response}
& 1 & Diagnostic verification, field case definitions and line lists
& Confirms the pathogen, applies field case definitions, and builds and cleans line lists. \\
& & 2 & Event specific attack rates and epidemic curves
& Quantifies spread in defined groups and interprets epidemic curves for the event. \\
& & 3 & Outbreak hypothesis mapping and source attribution
& Links cases by time, place, and shared exposures to identify likely sources and transmission chains, integrating line lists, environmental sampling, traceback, and genomic evidence. \\
& & 4 & Immediate control and situation reporting
& Implements urgent measures and documents current status with concise situation reports. \\

\bottomrule
\end{tabularx}
\label{tab:taxonomy_topics_12}
\end{table*}

\begin{table*}[t]
\centering
\caption{Epidemiology taxonomy topics (Classes 3 and 4).}
\label{tab:taxonomy_topics_34}
\footnotesize
\setlength{\tabcolsep}{5pt}
\renewcommand{\arraystretch}{1.12}
\begin{tabularx}{\textwidth}{L{0.04\textwidth} L{0.22\textwidth} L{0.04\textwidth} L{0.22\textwidth} X}
\toprule
\textbf{Cls} & \textbf{Class} & \textbf{Top} & \textbf{Topic} & \textbf{Description} \\
\midrule

\multirow[t]{4}{*}{3} & \multirow[t]{4}{=}{Determinants and Exposures}
& 1 & Contextual determinants of exposure
& Integrates individual behaviors with environmental, occupational, and social and structural conditions that shape exposure probability and inequities. \\
& & 2 & Transmission modes and contact patterns
& Describes general routes of spread and population contact structures across settings. \\
& & 3 & Exposure response interpretation
& Specifies the exposure metric, determines whether values are above or below assay limits and thresholds, and interprets exposure to infection, severity, or transmissibility patterns as reported in the passage. \\
& & 4 & Zoonotic and One Health interfaces, reservoirs and vectors
& Identifies animal reservoirs, vectors, and human animal environment interfaces where spillover can occur. \\
\addlinespace

\multirow[t]{4}{*}{4} & \multirow[t]{4}{=}{Susceptibility and Immunity}
& 1 & Susceptibility stratification and special populations
& Identifies groups more susceptible to infection based on demographic and clinical traits and setting specific contexts. \\
& & 2 & Serology and correlates of protection
& Estimates seroprevalence and relates immune markers to protection thresholds and population level immunity. \\
& & 3 & Protection effectiveness, waning, reinfection and immune escape
& Describes protection after vaccination or prior infection, its change over time, risks of reinfection, hybrid immunity, and variant related escape, considers how vaccine dose number and dose intervals influence vaccine effectiveness and its waning over time. \\
& & 4 & Severity risk and prognostic factors
& Assesses risk of severe outcomes conditional on infection and stratifies prognosis by host factors. \\

\bottomrule
\end{tabularx}
\end{table*}

\begin{table*}[t]
\centering
\caption{Epidemiology taxonomy topics (Classes 5 and 6)}
\label{tab:taxonomy_topics_56}
\footnotesize
\setlength{\tabcolsep}{5pt}
\renewcommand{\arraystretch}{1.12}
\begin{tabularx}{\textwidth}{L{0.04\textwidth} L{0.22\textwidth} L{0.04\textwidth} L{0.22\textwidth} X}
\toprule
\textbf{Cls} & \textbf{Class} & \textbf{Top} & \textbf{Topic} & \textbf{Description} \\
\midrule

\multirow[t]{5}{*}{5} & \multirow[t]{5}{=}{Modeling, Methods, and Evaluation}
& 1 & Transmission modeling and inference
& Uses mechanistic or statistical models to estimate transmission parameters and infer transmission patterns. \\
& & 2 & Real time debiasing and delay adjustment
& Reconstructs recent incidence by adjusting for reporting delays, right truncation, and under ascertainment. \\
& & 3 & Study design and causal effects
& Selects designs and identification strategies and defines effect measures for causal estimation. \\
& & 4 & Measurement and bias handling
& Addresses measurement validity, misclassification and measurement error, confounding and selection, generalizability, survey weighting, and sample size. \\
& & 5 & Program performance and impact evaluation
& Assesses coverage and implementation fidelity, audits routine data quality, evaluates real world effectiveness, and estimates disease burden. \\
\addlinespace

\multirow[t]{4}{*}{6} & \multirow[t]{4}{=}{Projections and Forecasts}
& 1 & Near term forecasting
& Produces short horizon probabilistic forecasts for upcoming values and quantifies forecast uncertainty. \\
& & 2 & Scenario projections
& Projects future trajectories under stated assumptions about policy, behavior, or immunity. \\
& & 3 & Forecast evaluation and model combination
& Assesses forecast quality using proper scoring rules, calibration, and sharpness diagnostics, and develops or applies methods to combine multiple forecasting models to improve predictive accuracy, stability, and robustness across contexts. \\
& & 4 & Decision oriented forecasting and risk communication
& Maps forecast probabilities to operational thresholds or cost loss trade offs and communicates uncertainty for decision making. \\

\bottomrule
\end{tabularx}
\end{table*}

\subsection{External Knowledge Construction}
\label{app:external_knowledge}

This appendix describes how external knowledge $\mathcal{E}$ is constructed for EpiQAL-B. The procedure consists of four steps: entity extraction, entity linking, triple retrieval, and summarization.

We first extract disease entities from the source document using GLiNER \citep{zaratiana-etal-2024-gliner}. Extracted mentions are then normalized via entity linking using SapBERT \citep{liu2021self}, which is a SOTA biomedical entity linking method \citep{xie2024promptlink}, to encode mentions and retrieve candidate entities. We retrieve related triples from two knowledge graphs: eKG-DONs \citep{JRC139580}, which compiles outbreak reports from official sources, and iBKH \citep{Su2021.03.12.21253461}, which encodes broader biomedical relations. Finally, a language model summarizes the retrieved triples into compact natural language statements used as generation signals \citep{xie2025kerapknowledgeenhancedreasoningapproach}.

These signals are used only during dataset construction to steer the generation model toward inference-oriented questions. They are not  provided to models at evaluation time.

\subsection{Distractor Design}
\label{app:distractors}

We design distractors to be plausible under the provided study context while remaining incorrect for the specific question intent. Across all subsets, we enforce semantic type matching with correct options, stylistic consistency, and diversity so that different distractors reflect different confusable alternatives rather than near duplicates. We attach evidence spans and brief rationales during construction to support verification and error analysis. Note that the design categories described here reflect construction intent. The post-hoc distractor classification used in the error analysis (\S\ref{sec:error}) is performed independently by GPT-5-mini and uses a different category scheme; see Appendix~\ref{app:distractor_analysis} for definitions.

\vskip 0.2em \noindent\textbf{EpiQAL-A.}
Distractors in EpiQAL-A are passage-grounded confounders. They are valid entities or facts stated in the same document, matching the semantic category and tone of correct options. They are incorrect because they refer to a different role, population, setting, time window, or study context than what the question requires. This design discourages guessing by surface cues while preserving a retrieval-based task in which all options are locally supported by explicit spans.

 \vskip 0.2em \noindent\textbf{EpiQAL-B.}
Distractors in EpiQAL-B are reasoning-level adversaries. They share the grammatical structure and semantic category of correct options but express misleading implications that require a reasoning process. We introduce subtle flaws using three main categories:
\begin{enumerate}[label=\textbullet, leftmargin=*, nosep]
    \item \textbf{Entity or attribution shift}: a conclusion that holds for another entity in the passage is incorrectly applied to the target entity.
    \item \textbf{Causal direction reversal}: the direction of an implied effect is flipped while keeping entities and study context fixed.
    \item \textbf{Principle mismatch}: a correct passage fact is combined with an incorrect epidemiological principle to yield a plausible but wrong implication.
\end{enumerate}
Construction-time external signals may validate the flawed reasoning chain but are not embedded as explicit hints in the distractor text.

\vskip 0.2em \noindent\textbf{EpiQAL-C.}
Distractors in EpiQAL-C are masked-input traps tailored to the Discussion masking setup. We draw candidates from either the non-Discussion sections or the Discussion, then refine them into self-contained sentences that are plausible but incorrect when only the non-Discussion sections are available. We use five primary trap categories:
\begin{enumerate}[label=\textbullet, leftmargin=*, nosep]
    \item \textbf{Limitations or future work}: unproven hypotheses that are not established as conclusions.
    \item \textbf{External literature dependence}: claims supported only by cited outside work in the Discussion.
    \item \textbf{Background restatement}: common knowledge rather than study-specific findings.
    \item \textbf{Incorrect conclusion}: same entity but wrong conclusion under the question.
    \item \textbf{Causal reversal}: reversed causal direction under the study context.
\end{enumerate}
For each distractor, we attach evidence revealing why it is not a valid answer under the masked-input setting.

\subsection{Stem Refinement Procedure \& Example}
\label{app:stem_refine_details}

Stem refinement is a retrieval-based rewriting step applied during dataset construction. We adapt the recursive retrieval approach from \citet{wu2025webdancer} by iteratively replacing entities with their descriptions.

The procedure works as follows. First, we prompt a model to extract a core entity from the question stem as a replacement candidate. Second, we construct a synthetic query to search for the entity's definition and characteristics, retrieving the top $K_r$ relevant snippets from the web. Third, a model summarizes these snippets into a concise description that replaces the original entity in the stem. This process repeats until the DiffScore exceeds threshold $\theta_d$ or reaches the maximum number of iterations $T_r$. No retrieved text is provided to models at evaluation time; only the rewritten stem is used.

Table~\ref{tab:refine_example} shows a representative instance before and after refinement. Refinement replaces salient entities with descriptive phrases that preserve answerability but remove direct lexical anchors. This requires models to map descriptions back to the correct concepts and integrate evidence from the passage.

\begin{table*}[t]
\centering
\small
\caption{An example of stem refinement. The options are unchanged, and only the question stem is rewritten.}
\label{tab:refine_example}
\begin{tabular}{p{0.12\linewidth}p{0.82\linewidth}}
\hline
\textbf{Version} & \textbf{Question stem} \\
\hline
Original & \textit{Which of the following best captures the primary implication of integrating patient-reported experiences and preferences into the early-stage development of medicinal products for neglected tropical diseases, based on the qualitative findings from a multi-country study on \underline{cutaneous leishmaniasis}?} \\
\\
Iteration 1 & \textit{Which of the following best captures the primary implication of integrating patient-reported experiences and preferences into the early-stage development of medicinal products for neglected tropical diseases, based on the qualitative findings from a multi-country study on \textbf{a vector-borne skin disorder caused by \underline{Leishmania parasites}, characterized by painless, chronic ulcers or nodules on exposed body parts, primarily resulting from sandfly bites and affecting millions globally}?} \\
\\
Iteration 2 & \textit{Which of the following best captures the primary implication of integrating patient-reported experiences and preferences into the early-stage development of medicinal products for \underline{neglected tropical diseases}, based on the qualitative findings from a multi-country study on \textbf{a vector-borne skin disorder caused by protozoan parasites from over 20 species transmitted to humans via bites of infected phlebotomine sandflies}, primarily causing chronic skin lesions through vector-borne transmission, affecting millions globally?} \\
\\
Iteration 3 & \textit{Which of the following best captures the primary implication of integrating patient-reported experiences and preferences into the early-stage development of medicinal products for \textbf{a diverse group of communicable diseases caused by parasitic, bacterial, fungal, viral, and protozoan pathogens, predominantly affecting impoverished populations in tropical and subtropical regions and perpetuating cycles of poor health, social marginalization, and economic hardship}, based on the qualitative findings from a multi-country study on a vector-borne skin disorder caused by protozoan parasites from over 20 species transmitted to humans via bites of infected phlebotomine sandflies, primarily causing chronic skin lesions through vector-borne transmission, affecting millions globally?} \\
\hline
\end{tabular}
\end{table*}

In Table~\ref{tab:refine_example}, \underline{underlined text} marks the entity selected for replacement at each iteration, and \textbf{bold text} indicates the retrieved description that replaces the original surface form. In Iteration 1, \underline{cutaneous leishmaniasis} is replaced with a descriptive paraphrase. Iteration 2 expands \underline{Leishmania parasites} into a higher-level description while preserving question intent. In Iteration 3, \underline{neglected tropical diseases} is replaced, further reducing lexical overlap between the stem and source evidence. To answer correctly, models must identify which epidemiological entity the description refers to and use passage evidence to select the correct options, rather than relying on surface-form matching.

\subsection{Difficulty Control Analysis}
\label{app:difficulty_analysis}
 
\begin{figure}[t]
    \centering
    \includegraphics[width=\columnwidth]{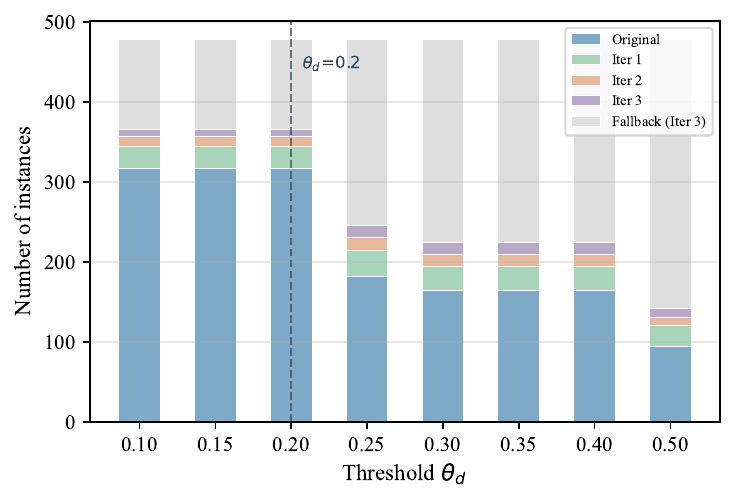}
    \\[0.3em]
    \includegraphics[width=\columnwidth]{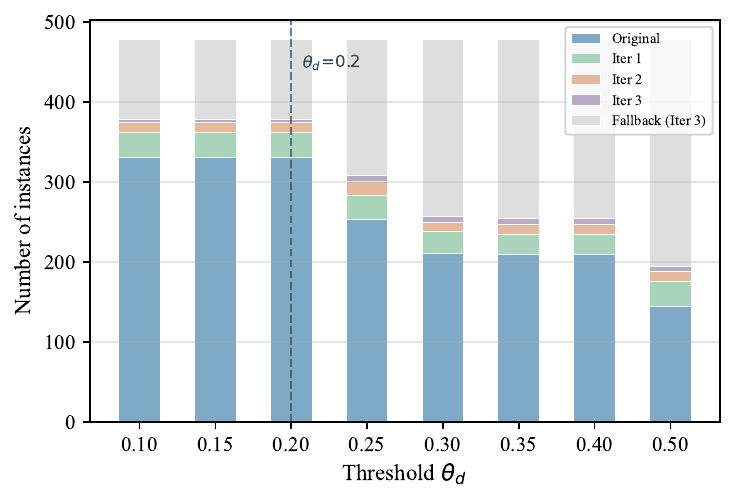}
    \caption{Version selection distribution across difficulty thresholds for EpiQAL-B (top) and EpiQAL-C (bottom). Each bar shows, for a given $\theta_d$, how many instances first meet the threshold at each iteration. The dashed line marks $\theta_d = 0.2$; raising it to 0.25 sharply increases the number of instances requiring refinement.}
    \label{fig:threshold_selection}
\end{figure}

\vskip 0.2em \noindent\textbf{Threshold selection.}
We set $\alpha = 0.3$ to weight EM more heavily than F1, reflecting our emphasis on exact set recovery over partial overlap. To select a threshold, we first apply all three refinement iterations to every instance regardless of difficulty, producing four complete DiffScore distributions (Original through Iter 3). Figure~\ref{fig:threshold_selection} shows how the version selection distribution changes across thresholds under this setup. A sharp transition occurs between $\theta_d = 0.20$ and $0.25$: the fraction of easy instances in EpiQAL-B jumps from 33.7\% to 61.9\%, and in EpiQAL-C from 30.9\% to 47.2\%, because the median DiffScore falls in this range. We therefore set $\theta_d = 0.2$, which aligns with a natural difficulty boundary and avoids excessive refinement. In the final pipeline, for each instance we select the first refinement iteration where DiffScore $\geq \theta_d$; if no iteration meets the threshold, the final iteration (Iter 3) is used as fallback.
 
\begin{figure*}[t]
    \centering
    \includegraphics[width=\columnwidth]{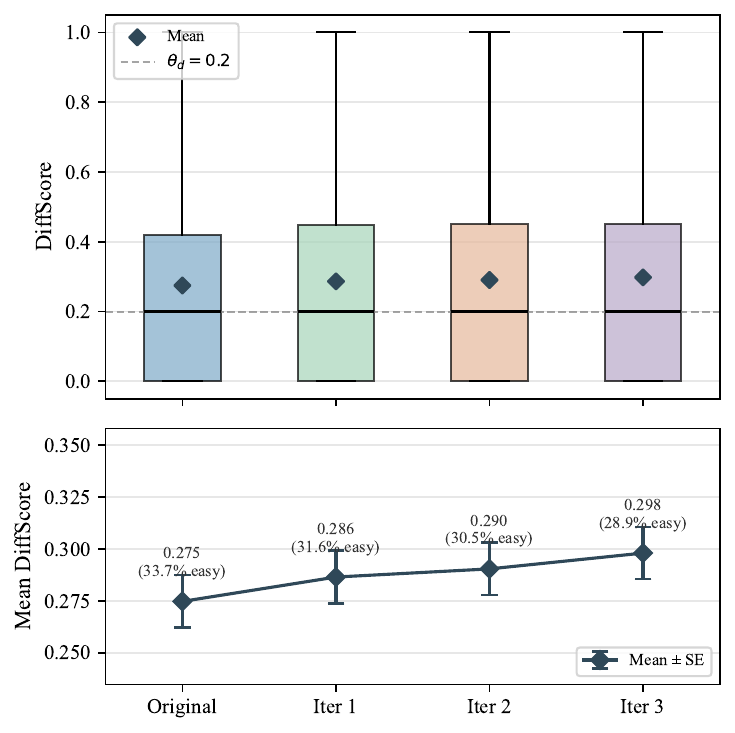}
    \includegraphics[width=\columnwidth]{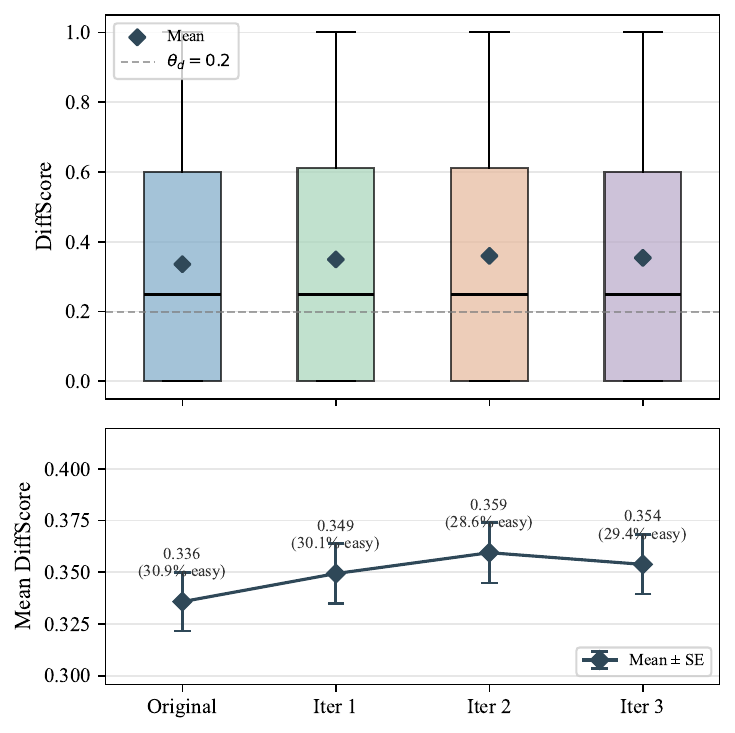}
    \caption{DiffScore distribution across stem refinement iterations for EpiQAL-B (left) and EpiQAL-C (right), computed by applying all iterations to every instance. Diamonds indicate means; the dashed line marks $\theta_d = 0.2$. The lower panels zoom into the mean trend with standard error bars.}
    \label{fig:diffscore_dist}
\end{figure*}

\vskip 0.2em \noindent\textbf{DiffScore distribution.}
To characterize how refinement affects difficulty, Figure~\ref{fig:diffscore_dist} shows the DiffScore distribution at each iteration when all instances are uniformly refined (i.e., without applying $\theta_d$ selection). This isolates the effect of stem rewriting from the selection process. Refinement produces a modest but consistent upward shift. For EpiQAL-B, the mean DiffScore increases from 0.275 (Original) to 0.298 (Iter 3), and the proportion of easy instances ($\text{DiffScore} < \theta_d$) drops from 33.7\% to 28.9\%. For EpiQAL-C, the mean increases from 0.336 to 0.354, with easy instances decreasing from 30.9\% to 29.4\%. The diminishing marginal gains across iterations motivate the choice of $T_r = 3$.
 
\vskip 0.2em \noindent\textbf{Final selection.}
After applying the $\theta_d = 0.2$ threshold, the pipeline selects the earliest iteration meeting the criterion for each instance. For EpiQAL-B, 317 instances use the original stem (already non-trivial), 28 use Iter 1, 12 use Iter 2, and 121 use Iter 3 (of which 112 are fallback cases where no iteration met $\theta_d$). For EpiQAL-C, the corresponding counts are 331, 31, 12, and 105 (of which 101 are fallback). The majority of instances already exceed the difficulty threshold before refinement, supporting the interpretation that the pipeline primarily targets genuinely easy items rather than rewriting indiscriminately.

\subsection{Construction Pipeline Pseudocode}
\label{app:pseudocode}
Algorithm~\ref{alg:pipeline} summarizes the EpiQAL construction pipeline. Subset-specific steps are marked with their applicable subsets.

\begin{algorithm*}[htbp]
\caption{EpiQAL Construction Pipeline}
\label{alg:pipeline}
\small
\begin{algorithmic}[1]
\Require Generator $\mathcal{M}_g$, checkers $\{C_j\}_{j=1}^3$, difficulty pool $\{P_j\}_{j=1}^k$, thresholds $\theta_h, \theta_c, \theta_d$, max refinement iterations $T_r$
\Statex
\For{each document $\mathcal{D}$ in corpus}
\Statex
\State \textbf{Input Processing.} \, \textbf{[A, B]} Classify question type for $\mathcal{D}$; choose taxonomy topic $\tau$. \textbf{[B]} Extract disease entities via NER; link to KG; summarize triples $\to \mathcal{E}$. \textbf{[C]} Separate $\mathcal{D}$ into body $\tilde{\mathcal{D}}$ and Discussion $\mathcal{D}_d$. Set $\mathcal{E} \gets \emptyset$ unless populated by~B.
\Statex
\State \textbf{QA Generation.} \, \textbf{[A, B]} Generate $\mathcal{Q}$ from $(\mathcal{D}, \tau, \mathcal{E})$, then $\mathcal{O}_c$ from $(\mathcal{D}, \mathcal{Q}, \mathcal{E})$, then $\mathcal{O}_d$ from $(\mathcal{D}, \mathcal{Q}, \mathcal{O}_c, \mathcal{E})$. \textbf{[C]} Extract $\mathcal{O}_c$ from $\mathcal{D}$ (sourced from $\mathcal{D}_d$), generate $\mathcal{Q}$ from $(\mathcal{D}, \mathcal{O}_c)$, then $\mathcal{O}_d$ from $(\mathcal{D}, \mathcal{Q}, \mathcal{O}_c)$.
\Statex
\State \textbf{Multi-model Verification.} \, For each $o \in \mathcal{O}_c \cup \mathcal{O}_d$: each checker $C_j$ votes 3 times (temp 1.0), yielding $v = \text{votes}/9$. Accept if $v \geq \theta_h$; reject if $v < \theta_c$; flag for human review otherwise. Discard instance if all $\mathcal{O}_c$ rejected.
\Statex
\State \textbf{Difficulty Control \textup{[B, C]}.} \, Evaluate $\{P_j\}$ on $(\tilde{\mathcal{D}}, \mathcal{Q}, \mathcal{O})$; compute DiffScore $= 1 - \frac{1}{k}\sum_j(\alpha \cdot F_1^{(j)} + (1{-}\alpha) \cdot \text{EM}^{(j)})$. While DiffScore $< \theta_d$ and $t \leq T_r$: extract salient entity from $\mathcal{Q}$, retrieve description, replace in $\mathcal{Q}$, recompute DiffScore.
\Statex
\State Output $(\tilde{\mathcal{D}}, \mathcal{Q}, \mathcal{O}_c, \mathcal{O}_d)$
\EndFor
\end{algorithmic}
\end{algorithm*}

\section{Experimental Details}
\label{app:exp_details}
 
\subsection{Compute and Inference Settings}
Experiments run on NVIDIA H100 and H200 GPUs. Llama-3.3-70B-Instruct and GLM-4.5-Air use four-bit inference, and all other models use default precision settings.
 
\subsection{Preprocessing}
We extract structured sections when available and normalize raw text by removing reference lists and non-content artifacts. Documents are assembled in a fixed section order to reduce variance across instances. We drop papers with missing main text or abnormal formatting that prevents reliable section parsing.
 
\subsection{Model Configuration}
 
\vskip 0.2em \noindent\textbf{Generation model.}
We use Qwen3-30B-A3B-Instruct-2507 as the generation model. For disease entity extraction, we use GLiNER \citep{zaratiana-etal-2024-gliner}. For entity linking in EpiQAL-B construction, we use SapBERT \citep{liu2021self} to encode mentions and retrieve candidate disease entities from knowledge graphs. To summarize knowledge graph triples into natural language signals, we also use Qwen3-30B-A3B-Instruct-2507. Generation temperature is set to 0 for reproducibility.
 
\vskip 0.2em \noindent\textbf{Checking model group.}
We verify generated options using three instruction-tuned models from different families: GPT-5-mini, DeepSeek-V3.2-Thinking, and GLM-4.5-Air. Each checker runs 3 times with temperature 1.0, yielding 9 total votes per option, and decisions are aggregated into the vote ratio $v$ defined in Section~\ref{sec:modelverify}. We set the acceptance threshold $\theta_h = 6/9$ and the rejection threshold $\theta_c = 5/9$: options with $v \geq \theta_h$ are accepted automatically, options with $v < \theta_c$ are rejected, and options with $\theta_c \leq v < \theta_h$ are flagged for human review.
 
\vskip 0.2em \noindent\textbf{Difficulty judging pool.}
To estimate difficulty as described in Section~\ref{sec:difficulty}, we evaluate a pool of four models spanning different capability tiers: GPT-5-mini, DeepSeek-V3.2-Thinking, Qwen3-32B, and Phi-4-mini-instruct. We compute DiffScore with $\alpha = 0.3$ and average across models. The difficulty threshold is $\theta_d = 0.2$, maximum refinement iterations $T_r = 3$, and retrieval budget $K_r = 6$ snippets.


\section{Dataset Analysis}
\label{app:dataset_analysis}

This appendix provides additional analysis of dataset composition for EpiQAL-A and EpiQAL-B, which use taxonomy-guided generation. EpiQAL-C derives supervision from paper structure rather than taxonomy and is not included in this analysis.

\subsection{Class and Topic Distribution}
\begin{figure*}[th!]
    \centering
    \includegraphics[width=0.9\textwidth, page=1, trim=0cm 0cm 0cm 0cm, clip]{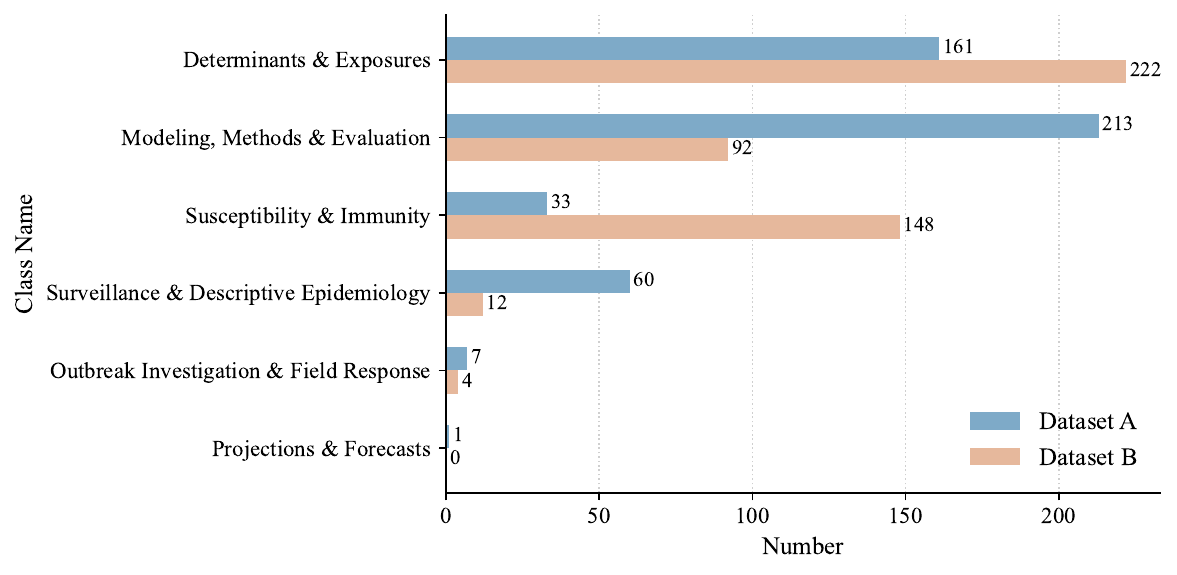}
    \caption{Class distribution for EpiQAL-A and EpiQAL-B.}
  \label{fig:classdis}
\end{figure*}

Figure~\ref{fig:classdis} shows the distribution of instances across the six taxonomy classes. Coverage is uneven across classes. Modeling, Methods and Evaluation is the most frequent class in EpiQAL-A (213 instances) and Determinants and Exposures in EpiQAL-B (222), while Outbreak Investigation and Field Response and Projections and Forecasts together account for fewer than 10 instances in either subset. This distribution reflects the prevalence of these topics in the source corpus of neglected tropical disease research.

\begin{figure*}[th!]
    \centering
    \includegraphics[width=\textwidth, page=1, trim=0.5cm 0.5cm 0.5cm 0cm, clip]{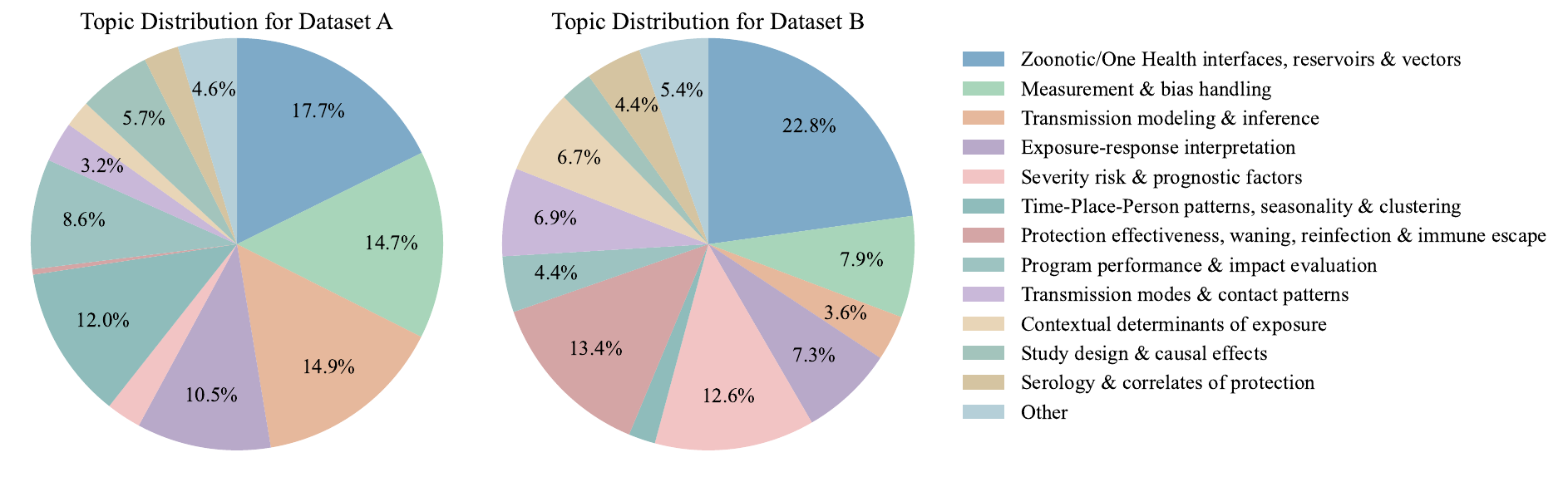}
    \caption{Topic distribution for EpiQAL-A and EpiQAL-B.}
  \label{fig:topicdis}
\end{figure*}

Figure~\ref{fig:topicdis} shows the most frequent topics, with the remainder grouped as Other. Zoonotic and One Health interfaces is the single most frequent topic in both subsets (17.7\% and 22.8\%), followed by measurement, modeling, and severity-related topics.

\subsection{Verification Example}
\label{app:checker_example}

Table~\ref{tab:checker_example} illustrates the checker's decision process on an instance flagged by a reviewer during the initial submission. The question asks \textit{``Which two Acanthamoeba species were used in the study to evaluate their interaction with Leishmania promastigotes?''} with three candidate options.

\begin{table}[htbp]
\caption{Checker example. The third option is a reordering of the correct answer; all nine votes reject it as an invalid distractor.}
\centering
\small
\begin{tabular}{lccl}
\toprule
\textbf{Option} & \textbf{Label} & \textbf{Vote} & \textbf{Dec.} \\
\midrule
A. castellanii \& A. polyphaga & Corr. & 9/9 & Acc. \\
A. castellanii \& A. spp. & Dist. & 9/9 & Acc. \\
A. polyphaga \& A. castellanii & Dist. & 0/9 & Rej. \\
\bottomrule
\end{tabular}
\label{tab:checker_example}
\end{table}

The third option reverses the order of the two species but is factually equivalent to the correct answer. All nine checker votes unanimously reject it as an invalid distractor, with rationales identifying that reordering does not constitute a substantive error. This case demonstrates that the upgraded checking group reliably detects semantic equivalence between superficially distinct options, addressing a limitation identified during initial review.

\section{Human Evaluation Details}
\label{app:human_eval}

\vskip 0.2em \noindent\textbf{Protocol.}
To assess the quality of the finalized benchmark, three computer science PhD students with biomedical training independently score a stratified sample of 120 questions (40 per subset). The 120-question sample represents 8.4\% of the benchmark, stratified to cover the full difficulty range. This evaluation is conducted after benchmark construction is complete and does not influence dataset content. Annotators may use LLMs to aid comprehension of domain-specific text, but all scoring is performed by human judgment. For EpiQAL-B and EpiQAL-C, we stratify by difficulty: 12 Easy (DiffScore = 0), 14 Medium (0 < DiffScore $\leq$ 0.4), and 14 Hard (DiffScore > 0.4). For EpiQAL-A, we use random sampling since difficulty screening is not applied. Annotators rate each question on subset-specific dimensions using a 1--3 scale (1 = poor, 2 = acceptable, 3 = good). We report Exact Agreement (percentage of items where all annotators assign the same score), Binary Agreement (collapsing to acceptable vs.\ poor), and Severe Disagreement (items where both 1 and 3 appear) for inter-annotator reliability.

\vskip 0.2em \noindent\textbf{Dimension-level Analysis.}
Answer Correctness scores highest across all subsets ($\geq$2.81), with EpiQAL-C at 2.92, indicating that generated answers reliably match the source passages. Evidence Sufficiency is also strong: EpiQAL-A scores 2.93 since its questions are grounded in single passages, while EpiQAL-B and C score 2.75 and 2.77 due to more complex evidence chains. Distractor Quality is comparable across subsets (2.52--2.56), suggesting that generating plausible yet incorrect alternatives is the hardest part of automated benchmark construction regardless of subset type. Question Clarity decreases from EpiQAL-A (2.77) to EpiQAL-C (2.51), as expected given the increasing complexity of question phrasing from retrieval to conclusion reconstruction. Reasoning Depth is 2.45 for both EpiQAL-B and C, indicating that most questions do require non-trivial inference. Answerability for EpiQAL-C (2.78) confirms that conclusion reconstruction questions are solvable from the provided evidence.

\vskip 0.2em \noindent\textbf{Inter-Annotator Agreement.}
Exact Agreement on the 3-point scale is 68.8\% for EpiQAL-A, 47.5\% for EpiQAL-B, and 55.4\% for EpiQAL-C. EpiQAL-B has the lowest agreement, which is expected given the subjectivity in judging reasoning depth and distractor plausibility for multi-step inference questions. However, these numbers reflect the narrow scoring range more than genuine disagreement: Binary Agreement exceeds 91\% on all subsets (A: 95.0\%, B: 91.5\%, C: 91.7\%), and Severe Disagreement is low (A: 3.1\%, B: 7.5\%, C: 5.0\%). Across subsets, 82--88\% of disagreements are between adjacent scores (2 vs.\ 3), meaning annotators largely agree on whether questions are acceptable and only differ on the degree of quality.

\vskip 0.2em \noindent\textbf{Expert Validation.}
To validate annotator reliability, a faculty member in biostatistics and epidemiology scored 30 questions (10 per subset) using the same dimensions and scale. This is the same faculty member who co-designed the taxonomy and is an author of this paper. Scoring was performed after benchmark construction was complete and blind to the annotators' scores, so the validation is independent of the annotation pass rather than of the project. We compare the expert's scores against the PhD annotators' majority vote on each dimension-item pair. Binary Agreement is 97.5\% (A, 40 pairs), 96.0\% (B, 50 pairs), and 98.3\% (C, 60 pairs). Severe Disagreement is 2.5\% (A), 2.0\% (B), and 0.0\% (C). Per-dimension breakdowns are not reported due to the small per-subset sample size (10 items), but no dimension shows Binary Agreement below 90\% on any subset.

\section{Additional Results}
\subsection{Stem Refinement Effect on Model Performance}
\label{app:stem_refine_analysis}
 
\begin{table}[t]
    \centering 
    \caption{Exact Match accuracy on EpiQAL-C across stem refinement iterations (w/o CoT). Models are the difficulty judging pool. \textbf{Bold} = highest, \underline{underline} = lowest per row. Ties are both marked.}
    \label{tab:refine}
    \small
    \begin{tabular}{lcccc}
        \toprule
         \textbf{Diff. Judging Model} & \textbf{Orig.} & \textbf{Iter 1} & \textbf{Iter 2} & \textbf{Iter 3} \\
        \midrule
        GPT-5-mini          & \underline{0.595} & 0.599 & \underline{0.595} & \textbf{0.601} \\
        DeepSeek-V3.2   & \textbf{0.727} & 0.701 & \underline{0.697} & 0.704 \\
        Qwen3-32B           & \textbf{0.686} & 0.662 & 0.635 & \underline{0.630} \\
        Phi-4-mini          & \textbf{0.446} & \underline{0.411} & 0.415 & 0.425 \\
        \midrule
        \textbf{Average}             & \textbf{0.613} & 0.593 & \underline{0.586} & 0.590 \\
        \bottomrule
    \end{tabular}
\end{table}
 
To isolate the effect of refinement, we construct controlled variants of EpiQAL-C by applying 0 to $T_r$ refinement iterations to the same base instances, regardless of whether they would be refined in the final pipeline. We evaluate each model in the difficulty judging pool without Chain-of-Thought prompting at temperature 0. Results are shown in Table~\ref{tab:refine}.
 
As shown in Table~\ref{tab:refine}, average EM decreases from 0.613 (Original) to 0.586 (Iter 2), confirming that stem refinement increases difficulty. The maximum decline across iterations is most pronounced for Qwen3-32B (-0.056, at Iter 3) and Phi-4-mini (-0.035, at Iter 1), while GPT-5-mini shows minimal sensitivity. Interestingly, Iter 3 slightly recovers compared to Iter 2 for three of four models.  We attribute this to over-expansion: by the third iteration, entity descriptions become sufficiently detailed (see Table~\ref{tab:refine_example}) that they inadvertently supply domain knowledge, partially compensating for the loss of direct lexical anchors. Considering the trade-off between generation efficiency and difficulty gain, we set $T_r=3$.

\subsection{Distractor Analysis Details}
\label{app:distractor_analysis}

\vskip 0.2em \noindent\textbf{Category definitions.}
For EpiQAL-A: \textit{Wrong entity/role} (correct fact applied to a different entity or misattributed role), \textit{Wrong context} (correct fact from a different group, location, time, or condition), \textit{Wrong metric} (correct fact but different measurement or statistic), \textit{Semantic near-miss} (similar wording but differs in a critical detail).

For EpiQAL-B: \textit{Methodological mismatch} (incompatible analysis method or study design), \textit{Causal/logical error} (reversed causation or broken reasoning chain), \textit{Variable confusion} (target variable confused with a related concept), \textit{Assumption violation} (correct terminology but violated underlying assumption).

For EpiQAL-C: \textit{External dependency} (requires information not in the Passage Body), \textit{Speculation/limitation} (untested hypothesis or methodological caveat), \textit{Background only} (general knowledge without study-specific interpretation), \textit{Causal reversal} (misinterpreted cause-effect from the Passage Body).

\vskip 0.2em \noindent\textbf{Deception rate.}
Let $C_c$ denote the set of distractors in category $c$, $M$ the set of evaluated models ($|M| = 14$; DeepSeek-V4-Flash-Thinking was not yet available when this analysis was run), and $e_{m,d} = 1$ if model $m$ misselects distractor $d$. The deception rate is:
$$\text{Deception Rate}(c) = \frac{\sum_{m \in M} \sum_{d \in C_c} e_{m,d}}{|C_c| \times |M|}$$

\vskip 0.2em \noindent\textbf{Distractor distribution.}
Table~\ref{tab:distractor_dist} reports the category distribution for each subset.

\begin{table}[h]
\centering
\caption{Distractor category distribution by subset.}
\label{tab:distractor_dist}
\small
\begin{tabular}{lcc}
\toprule
\textbf{Category} & \textbf{Count} & \textbf{\%} \\
\midrule
\multicolumn{3}{l}{\textit{EpiQAL-A} (n=960)} \\
\quad Wrong entity/role & 403 & 42.0\% \\
\quad Semantic near-miss & 242 & 25.2\% \\
\quad Wrong context & 203 & 21.1\% \\
\quad Wrong metric & 112 & 11.7\% \\
\midrule
\multicolumn{3}{l}{\textit{EpiQAL-B} (n=1,866)} \\
\quad Causal/logical error & 982 & 52.6\% \\
\quad Variable confusion & 689 & 36.9\% \\
\quad Assumption violation & 138 & 7.4\% \\
\quad Methodological mismatch & 57 & 3.1\% \\
\midrule
\multicolumn{3}{l}{\textit{EpiQAL-C} (n=1,787)} \\
\quad Speculation/limitation & 1,109 & 62.1\% \\
\quad External dependency & 379 & 21.2\% \\
\quad Causal reversal & 291 & 16.3\% \\
\quad Background only & 8 & 0.4\% \\
\bottomrule
\end{tabular}
\end{table}

\subsection{Per-Model Deception Rate}
\label{app:permodeldecrate}
 
Table~\ref{tab:per_model_deception} reports per-model overall deception rates on each subset under zero-shot no-CoT, along with the most vulnerable distractor category for each model. Models are sorted by average deception rate across subsets (ascending). Full per-category breakdowns are available in the released code repository.
 
\begin{table*}[htbp]
\centering
\small
\caption{Per-model overall deception rate (\%) and most vulnerable distractor category on each subset (zero-shot, no-CoT). Models are sorted by average rate across subsets.}
\label{tab:per_model_deception}
\begin{tabular}{lccccccr}
\toprule
 & \multicolumn{2}{c}{\textbf{EpiQAL-A}} & \multicolumn{2}{c}{\textbf{EpiQAL-B}} & \multicolumn{2}{c}{\textbf{EpiQAL-C}$^\dagger$} & \\
\cmidrule(lr){2-3} \cmidrule(lr){4-5} \cmidrule(lr){6-7}
\textbf{Model} & Rate & Top Cat. & Rate & Top Cat. & Rate & Top Cat. & \textbf{Avg.} \\
\midrule
DeepSeek-V3.2-Thinking   & 4.7  & Sem. n-m    & 6.1  & Var. conf.     & 5.8  & Spec./lim.  & 5.5 \\
Mistral-7B          & 8.4  & Sem. n-m    & 5.9  & Causal/log.    & 5.5  & Ext. dep.   & 6.6 \\
Mistral-Large       & 6.1  & Sem. n-m    & 12.0 & Var. conf.     & 9.3  & Spec./lim.  & 9.1 \\
GLM-4.5-Air         & 6.8  & Sem. n-m    & 12.2 & Var. conf.     & 12.5 & Spec./lim.  & 10.5 \\
GPT-4.1-nano        & 7.8  & Sem. n-m    & 10.7 & Assump. viol.  & 14.2 & Spec./lim.  & 10.9 \\
Qwen3-32B           & 7.5  & Sem. n-m    & 10.2 & Assump. viol.  & 18.7 & Spec./lim.  & 12.1 \\
GPT-5-mini          & 6.8  & Sem. n-m    & 18.5 & Var. conf.     & 11.7 & Spec./lim.  & 12.3 \\
Qwen3-30B-A3B       & 6.9  & Sem. n-m    & 17.3 & Var. conf.     & 15.4 & Spec./lim.  & 13.2 \\
Llama-3.3-70B       & 10.8 & Sem. n-m    & 14.2 & Var. conf.     & 16.1 & Spec./lim.  & 13.7 \\
Qwen3-8B            & 11.6 & Sem. n-m    & 22.3 & Var. conf.     & 25.9 & Spec./lim.  & 19.9 \\
Phi-4-mini          & 23.8 & Sem. n-m    & 27.1 & Meth. mism.    & 18.6 & Spec./lim.  & 23.2 \\
GPT-4o-mini         & 13.3 & Sem. n-m    & 34.4 & Var. conf.     & 31.7 & Spec./lim.  & 26.5 \\
Llama-3.2-3B        & 21.7 & Wrng. ctxt  & 31.0 & Causal/log.    & 28.9 & Spec./lim.  & 27.2 \\
Llama-3.1-8B        & 11.7 & Sem. n-m    & 32.0 & Causal/log.    & 40.9 & Spec./lim.  & 28.2 \\
\bottomrule
\end{tabular}

\vspace{0.5em}
\footnotesize
\noindent Sem.~n-m = Semantic near-miss, Var.~conf. = Variable confusion, Causal/log. = Causal/logical error, Meth.~mism. = Methodological mismatch, Assump.~viol. = Assumption violation, Spec./lim. = Speculation/limitation, Ext.~dep. = External dependency, Wrng.~ctxt = Wrong context.
\end{table*}

\renewcommand{\thefootnote}{$\dagger$}
\footnotetext{Background only (8 distractors) is excluded from the top-category comparison for EpiQAL-C.}
\renewcommand{\thefootnote}{\arabic{footnote}}

On EpiQAL-A, Semantic near-miss is the top vulnerability for 13 of 14 models, confirming it as the dominant failure mode regardless of model capability. On EpiQAL-B, Variable confusion is the most common top category (8 of 14 models), followed by Causal/logical error (3 models). On EpiQAL-C, Speculation/limitation dominates for all models except Mistral-7B, whose top category is External dependency, consistent with its flat deception profile noted in \S\ref{sec:discussion}.

\subsection{One-Shot Results}
\label{app:1shot}

Table~\ref{tab:1shot} reports Exact Match and F1 under one-shot no-CoT and CoT settings for 14 models using a single in-domain demonstration (excluded from evaluation).

\begin{table*}[htbp]
    \centering 
    \small
    \caption{One-shot Exact Match|F1 Score for each model across subsets, with and without Chain-of-Thought prompting. \textbf{Bold} = best, \underline{underline} = second best per column. Ties share the same mark.}
    \label{tab:1shot}
    \begin{tabular}{lcccccc}
        \toprule
         & \multicolumn{2}{c}{\textbf{EpiQAL-A}} & \multicolumn{2}{c}{\textbf{EpiQAL-B}} & \multicolumn{2}{c}{\textbf{EpiQAL-C}} \\
        \cmidrule(lr){2-3} \cmidrule(lr){4-5} \cmidrule(lr){6-7}
        \textbf{Model} & w/o CoT & CoT & w/o CoT & CoT & w/o CoT & CoT \\
        \midrule
        \textit{OpenAI} &&&&&&\\
        \hspace*{1em}GPT-5-mini                     & \underline{0.905}|\underline{0.966} & 0.916|0.963 & 0.553|0.802 & 0.654|0.836 & 0.579|0.787 & 0.577|0.774 \\
        \hspace*{1em}GPT-4o-mini                    & 0.773|0.910 & 0.802|0.915 & 0.488|0.762 & 0.501|0.772 & 0.289|0.695 & 0.297|0.695 \\
        \hspace*{1em}GPT-4.1-nano                   & 0.779|0.862 & 0.804|0.888 & 0.753|0.826 & 0.776|0.820 & 0.703|0.842 & 0.703|0.842 \\
        \midrule
        \textit{DeepSeek} &&&&&&\\
        \hspace*{1em}DeepSeek-V3.2-Thinking               & \textbf{0.928}|\textbf{0.970} & \textbf{0.926}|\textbf{0.970} & \textbf{0.841}|\textbf{0.903} & \textbf{0.895}|\textbf{0.914} & \underline{0.757}|0.842 & \underline{0.789}|0.854 \\
        \midrule
        \textit{Microsoft} &&&&&&\\
        \hspace*{1em}Phi-4-mini-instruct             & 0.583|0.811 & 0.730|0.857 & 0.342|0.720 & 0.532|0.763 & 0.513|0.774 & 0.506|0.776 \\
        \midrule
        \textit{Meta-Llama} &&&&&&\\
        \hspace*{1em}Llama-3.2-3B-Instruct           & 0.366|0.553 & 0.690|0.782 & 0.551|0.690 & 0.621|0.649 & 0.366|0.571 & 0.473|0.586 \\
        \hspace*{1em}Llama-3.1-8B-Instruct           & 0.798|0.911 & 0.827|0.905 & 0.616|0.804 & 0.828|0.869 & 0.232|0.641 & 0.418|0.730 \\
        \hspace*{1em}Llama-3.3-70B-Instruct          & 0.779|0.884 & 0.679|0.763 & 0.832|0.884 & \underline{0.876}|0.894 & 0.686|0.841 & 0.759|\textbf{0.865} \\
        \midrule
        \textit{Mistral AI} &&&&&&\\
        \hspace*{1em}Mistral-7B-Instruct-v0.3        & 0.722|0.809 & 0.745|0.828 & 0.790|0.812 & 0.830|0.830 & \textbf{0.808}|0.830 & \textbf{0.828}|0.831 \\
        \hspace*{1em}Mistral-Large-Instruct-2411     & 0.901|0.956 & \underline{0.922}|0.961 & 0.686|0.849 & 0.730|0.859 & 0.692|\textbf{0.856} & 0.678|0.854 \\
        \midrule
        \textit{Qwen} &&&&&&\\
        \hspace*{1em}Qwen3-8B                        & 0.811|0.921 & 0.863|0.940 & 0.585|0.809 & 0.690|0.843 & 0.611|0.814 & 0.657|0.830 \\
        \hspace*{1em}Qwen3-30B-A3B-Instruct-2507     & 0.882|0.956 & 0.911|\underline{0.965} & 0.767|0.878 & 0.864|\underline{0.906} & 0.665|0.838 & 0.734|\underline{0.862} \\
        \hspace*{1em}Qwen3-32B                       & 0.872|0.949 & 0.895|0.961 & \underline{0.841}|\underline{0.903} & 0.853|0.903 & 0.707|\underline{0.856} & 0.692|0.846 \\
        \midrule
        \textit{Zhipu AI} &&&&&&\\
        \hspace*{1em}GLM-4.5-Air                     & 0.874|0.947 & 0.903|0.962 & 0.751|0.884 & 0.780|0.886 & 0.669|0.849 & 0.686|0.854 \\
        \bottomrule
    \end{tabular}
\end{table*}

Compared with zero-shot results (Table~\ref{tab:bench}), one-shot prompting yields the most consistent gains on EpiQAL-B: the 14 models improve in EM without CoT, with Llama-3.2-3B showing the largest gain ($+$0.394) and Llama-3.1-8B improving by $+$0.298. On EpiQAL-A and EpiQAL-C, one-shot effects are smaller and less uniform. Notably, one-shot combined with CoT degrades Llama-3.3-70B on EpiQAL-A (0.789 $\to$ 0.679), likely due to the additional context length from the demonstration compounding with the full-article input.

\subsection{Role and Limitations of Retrieved Knowledge in EpiQAL-B}
\label{app:rag_analysis}

EpiQAL-B incorporates retrieval-augmented generation during benchmark construction: external triples from two biomedical knowledge graphs (eKG-DONs and iBKH) are provided as optional signals to encourage inference-oriented question design. We conducted a post-hoc analysis of all 478 finalized EpiQAL-B instances to assess whether and how retrieved knowledge contributed to the final questions.

Of the 478 instances, 27 (5.6\%) incorporated retrieved knowledge into the question or option design; the remaining instances relied solely on passage evidence. We manually inspected all 27 instances and found that in every case (27/27), the external knowledge provided only surface-level information that could not be integrated into the reasoning chain. Specifically, the retrieved triples restated facts already covered by the source passage or offered generic epidemiological context such as global disease distribution, without contributing any reasoning step that the passage evidence alone could not support. In no instance did external knowledge introduce a novel inferential link required to arrive at the correct answer. Table~\ref{tab:rag_example} illustrates a representative case.

\begin{table}[t]
\centering
\small
\caption{Representative example of retrieved knowledge in EpiQAL-B. Although external knowledge was formally incorporated during construction, it provides only surface-level context and does not contribute to the reasoning chain.}
\label{tab:rag_example}
\begin{tabular}{p{0.95\columnwidth}}
\toprule
\textbf{Question:} In a region where multiple vector-borne diseases are endemic, how might the spatial clustering of one disease inform the likely transmission dynamics of another, particularly when both are associated with human activity near international seaports? \\
\midrule
\textbf{Passage evidence:} (1) The risk of murine typhus was negatively associated with rainfall and temperature, after controlling for distance to in-use international seaports. (2) Kaohsiung is a hotspot for murine typhus, dengue, and scrub typhus, all vector-borne diseases common in this port city. (3) International seaports have long been hotspots for disease invasion due to expanding trade and international networks. \\
\midrule
\textbf{External knowledge:} Dengue Fever has been documented in numerous outbreaks across Asia, Africa, South America, and the Caribbean, with major events in Malaysia, India, Brazil, Pakistan, and Bangladesh. \\
\midrule
\textbf{Analysis:} The three passage evidence pieces already establish the full reasoning chain. Seaports facilitate pathogen introduction (Evidence 3), Kaohsiung is a port city where multiple vector-borne diseases co-occur despite involving different vectors (Evidence 2), and local climate modulates but does not override the port proximity effect (Evidence 1). The external knowledge that dengue has a broad global distribution adds no inferential step to this chain. The passage already states that dengue is prevalent in Kaohsiung; knowing that it also occurs in Brazil or Pakistan does not alter the reasoning about shared transmission dynamics in port environments. Removing the external knowledge leaves the question, the correct answer, and the underlying reasoning fully intact. \\
\bottomrule
\end{tabular}
\end{table}

We attribute this outcome to a granularity mismatch between the knowledge graphs and the source documents, which manifests in two ways.

\vskip 0.2em \noindent\textbf{High information density of source documents.}
Full-text research papers in PLOS Neglected Tropical Diseases contain detailed methodology, cross-referenced findings, and extensive background sections. This level of information density leaves minimal gaps for external knowledge to fill. Unlike short clinical queries or sparse patient notes, where external triples can supply missing context, long-form domain articles already embed sufficient evidence for multi-step inference.

\vskip 0.2em \noindent\textbf{Coarse granularity of general-purpose knowledge graphs.}
The triples in eKG-DONs and iBKH represent entity-level facts such as ``Dengue Fever, reported in, Asia, Brazil, Pakistan,'' which operate at a fundamentally different level of specificity than the study-level evidence in a research paper, where controlled variables, cohort characteristics, and measured effect sizes are situated within a specific geographic setting. This structural mismatch causes retrieved triples to remain at the surface level, unable to participate in the logical chains that connect study-specific observations to inferential conclusions.

\vskip 0.2em \noindent\textbf{Implications.}
These findings suggest that the utility of knowledge graph augmentation in benchmark construction is inversely related to the information density of the source document. When source texts are information-rich, general-purpose knowledge graphs degrade from an augmentation signal into a corroboration signal that confirms what the passage already states rather than extending the reasoning scope. Future work on knowledge-augmented QA construction may benefit from either domain-specialized knowledge resources with study-level granularity or source documents with sparser informational content such as clinical notes or public health bulletins, where external knowledge can address genuine evidence gaps.

\subsection{Question-Only Baseline}
\label{app:question_only}

To test whether models can answer EpiQAL questions without document evidence, we run a question-only baseline where all passage input is removed. Table~\ref{tab:question_only} reports Exact Match under zero-shot no-CoT.

\begin{table*}[htbp]
\centering
\small
\caption{Question-only baseline: Exact Match with full document input vs.\ question-and-options only (zero-shot, no-CoT).}
\label{tab:question_only}
\begin{tabular}{lcccccc}
\toprule
& \multicolumn{2}{c}{\textbf{EpiQAL-A}} & \multicolumn{2}{c}{\textbf{EpiQAL-B}} & \multicolumn{2}{c}{\textbf{EpiQAL-C}} \\
\cmidrule(lr){2-3} \cmidrule(lr){4-5} \cmidrule(lr){6-7}
\textbf{Model} & Full & Q-Only & Full & Q-Only & Full & Q-Only \\
\midrule
DeepSeek-V4-Flash-Thinking$^\ddagger$ & 0.926 & 0.023 & 0.703 & 0.538 & 0.666 & 0.035 \\
GPT-5-mini & 0.905 & 0.015 & 0.533 & 0.477 & 0.599 & 0.288 \\
GPT-4o-mini & 0.766 & 0.074 & 0.222 & 0.176 & 0.213 & 0.102 \\
GPT-4.1-nano & 0.768 & 0.131 & 0.678 & 0.431 & 0.559 & 0.466 \\
Phi-4-mini-instruct & 0.583 & 0.408 & 0.249 & 0.297 & 0.426 & 0.432 \\
Llama-3.2-3B-Instruct & 0.366 & 0.088 & 0.157 & 0.073 & 0.088 & 0.027 \\
Llama-3.1-8B-Instruct & 0.798 & 0.211 & 0.318 & 0.299 & 0.190 & 0.299 \\
Llama-3.3-70B-Instruct & 0.779 & 0.276 & 0.651 & 0.358 & 0.580 & 0.438 \\
Mistral-7B-Instruct-v0.3 & 0.722 & 0.297 & 0.789 & 0.418 & 0.808 & 0.559 \\
Mistral-Large-Instruct-2411 & 0.901 & 0.040 & 0.644 & 0.347 & 0.685 & 0.056 \\
Qwen3-8B & 0.811 & 0.038 & 0.508 & 0.368 & 0.484 & 0.278 \\
Qwen3-30B-A3B-Instruct-2507 & 0.882 & 0.076 & 0.573 & 0.500 & 0.585 & 0.175 \\
Qwen3-32B & 0.872 & 0.000 & 0.743 & 0.481 & 0.547 & 0.069 \\
GLM-4.5-Air & 0.874 & 0.021 & 0.657 & 0.364 & 0.580 & 0.104 \\
\bottomrule
\end{tabular}
\end{table*}

\renewcommand{\thefootnote}{$\ddagger$}
\footnotetext{DeepSeek-V4-Flash-Thinking replaces V3.2-Thinking as the latter is no longer available via API.}
\renewcommand{\thefootnote}{\arabic{footnote}}

On EpiQAL-A, average EM drops from 0.78 to 0.12, with several strong models (Qwen3-32B, GPT-5-mini, GLM-4.5-Air) falling to near zero, indicating that factual recall questions require document evidence. On EpiQAL-B and EpiQAL-C, residual question-only performance is higher because multi-choice options inherently carry partial information. To verify that stem refinement does not inflate question-only performance through descriptive entity replacements, we re-run the question-only setting on original (pre-refinement) stems for 13 models (excluding DeepSeek due to the version change). Average question-only EM is comparable with and without refinement (EpiQAL-B: 0.353 vs.\ 0.356; EpiQAL-C: 0.253 vs.\ 0.246), indicating that residual performance is not an artifact of refinement. Two models score no higher with the full document than without it on EpiQAL-C: Phi-4-mini (0.426 vs. 0.432) and Llama-3.1-8B (0.190 vs. 0.299). Both show large F1–EM gaps on that subset in Table \ref{tab:bench} (0.329 and 0.402), so this pattern is consistent with over-selection under long-context input penalizing Exact Match, rather than with these models ignoring the document. We note it as a case where Exact Match alone understates the contribution of the passage. Appendix~\ref{app:input_ablation} examines what this residual performance reflects.

\subsection{Input Ablation Controls}
\label{app:input_ablation}

To test how much of model performance depends on the source article, we compare four input conditions for the four API models. \textbf{Full} provides the correct article; \textbf{Shuffled} substitutes a randomly drawn article from the same subset; \textbf{Q-only} provides the question and options with no article (Table~\ref{tab:question_only}); \textbf{Option-only} provides the options alone. \textbf{Chance} is the empirical Exact Match of selecting one option at random. For EpiQAL-A this is a loose single-pick reference rather than a strict chance level, since EpiQAL-A admits multiple correct options and Exact Match requires recovering the full set.

\begin{table}[h]
\centering
\small
\setlength{\tabcolsep}{4pt}
\caption{Input ablation: Exact Match ranges over the four API models (zero-shot, no-CoT).}
\label{tab:input_ablation}
\begin{tabular}{lccccc}
\toprule
\textbf{Subset} & \textbf{Full} & \textbf{Shuf.} & \textbf{Q-only} & \textbf{Opt.} & \textbf{Chance} \\
\midrule
A & .77--.91 & $\sim$0 & .02--.13 & $\sim$0 & .31 \\
B & .23--.73 & .13--.45 & .18--.54 & .00--.17 & .21 \\
C & .20--.66 & .00--.13 & .04--.47 & .00--.07 & .22 \\
\bottomrule
\end{tabular}
\end{table}

On EpiQAL-A and EpiQAL-C, substituting a wrong article drops accuracy to near zero, and 82--100\% of answers are returned empty rather than incorrect. Shuffled also falls below Q-only, so a wrong passage is more harmful than no passage, which is consistent with models conditioning on the article. Option-only does not exceed the random single-option reference on any subset, so the options alone are not self-revealing; this does not rule out question-conditioned option elimination, which uses the question and options together and which these controls cannot isolate.

EpiQAL-B is the mixed case. Table~\ref{tab:input_ablation_b} breaks it out per model.

\begin{table}[h]
\centering
\small
\caption{EpiQAL-B input ablation, per model (Exact Match, zero-shot, no-CoT).}
\label{tab:input_ablation_b}
\begin{tabular}{lcccc}
\toprule
\textbf{Model} & \textbf{Full} & \textbf{Shuf.} & \textbf{Q-only} & \textbf{$\Delta$} \\
\midrule
GPT-5-mini    & 0.502 & 0.454 & 0.477 & $+$0.025 \\
DeepSeek-V4-Flash & 0.734 & 0.264 & 0.538 & $+$0.196 \\
GPT-4o-mini   & 0.230 & 0.130 & 0.176 & $+$0.054 \\
GPT-4.1-nano  & 0.671 & 0.347 & 0.431 & $+$0.240 \\
\bottomrule
\end{tabular}
\end{table}

DeepSeek-V4-Flash-Thinking and GPT-4.1-nano gain 0.196 and 0.240 EM from the correct passage relative to Q-only, whereas GPT-5-mini and GPT-4o-mini show little measurable benefit. Residual question-only performance is consistent with question-option informativeness, parametric knowledge, or a combination; the present controls do not separate these explanations, and we treat this as a remaining limitation rather than evidence for any single mechanism. The accuracies reported here come from re-running the API models and differ slightly from Table~\ref{tab:bench}, which we attribute to stochasticity and API variation.

\subsection{Temperature Sensitivity}
\label{app:temp_sensitivity}

To verify that our choice of temperature 0.3 does not affect benchmark conclusions, we re-run all evaluations at temperature 0 for the 12 models whose APIs support deterministic decoding (GPT-5-mini, DeepSeek-V3.2-Thinking, and DeepSeek-V4-Flash-Thinking are excluded). Table~\ref{tab:temp} reports Exact Match under zero-shot no-CoT.

\begin{table}[htbp]
\centering
\small
\setlength{\tabcolsep}{3pt}
\caption{Temperature sensitivity: Exact Match at T=0.3 vs.\ T=0 (zero-shot, no-CoT).}
\label{tab:temp}
\begin{tabular}{lcccccc}
\toprule
& \multicolumn{2}{c}{\textbf{EpiQAL-A}} & \multicolumn{2}{c}{\textbf{EpiQAL-B}} & \multicolumn{2}{c}{\textbf{EpiQAL-C}} \\
\cmidrule(lr){2-3} \cmidrule(lr){4-5} \cmidrule(lr){6-7}
\textbf{Model} & .3 & 0 & .3 & 0 & .3 & 0 \\
\midrule
GPT-4o-mini & .766 & .771 & .222 & .228 & .213 & .213 \\
GPT-4.1-nano & .768 & .775 & .678 & .661 & .559 & .553 \\
Phi-4-mini & .583 & .583 & .249 & .234 & .426 & .418 \\
Llama-3.2-3B & .366 & .383 & .157 & .146 & .088 & .088 \\
Llama-3.1-8B & .798 & .798 & .318 & .322 & .190 & .209 \\
Llama-3.3-70B & .779 & .777 & .651 & .653 & .580 & .578 \\
Mistral-7B & .722 & .726 & .789 & .808 & .808 & .810 \\
Mistral-Large & .901 & .899 & .644 & .626 & .685 & .681 \\
Qwen3-8B & .811 & .802 & .508 & .515 & .484 & .509 \\
Qwen3-30B-A3B & .882 & .878 & .573 & .573 & .585 & .587 \\
Qwen3-32B & .872 & .872 & .743 & .734 & .547 & .532 \\
GLM-4.5-Air & .874 & .880 & .657 & .655 & .580 & .587 \\
\bottomrule
\end{tabular}
\end{table}

Mean per-model EM deviation is below 1\% on every subset. Spearman rank correlation between T=0.3 and T=0 rankings exceeds 0.99 on all subsets (A: 0.993, B: 1.000, C: 0.995), confirming that the choice of temperature does not affect model rankings or benchmark conclusions.

\subsection{Multi-Run Stability}
\label{app:multi_run}

To verify result stability, we conduct three runs at the same settings for the 14 models. Table~\ref{tab:multi_run} reports mean and standard deviation of Exact Match under zero-shot no-CoT and CoT. Results are highly stable: the mean no-CoT standard deviation is below 1\% on each subset, while CoT shows slightly more variation but model rankings remain unchanged across runs.

\begin{table*}[htbp]
\centering
\small
\caption{Multi-run stability: mean$\pm$std of Exact Match (EM) across runs.}
\label{tab:multi_run}
\begin{tabular}{lcccccc}
\toprule
& \multicolumn{2}{c}{\textbf{EpiQAL-A}} & \multicolumn{2}{c}{\textbf{EpiQAL-B}} & \multicolumn{2}{c}{\textbf{EpiQAL-C}} \\
\cmidrule(lr){2-3} \cmidrule(lr){4-5} \cmidrule(lr){6-7}
\textbf{Model} & w/o CoT & CoT & w/o CoT & CoT & w/o CoT & CoT \\
\midrule
DeepSeek-V4$^\ddagger$ & 0.929$\pm$0.007 & 0.931$\pm$0.006 & 0.705$\pm$0.016 & 0.802$\pm$0.009 & 0.666$\pm$0.007 & 0.683$\pm$0.012 \\
GPT-5-mini & 0.898$\pm$0.005 & 0.927$\pm$0.002 & 0.531$\pm$0.002 & 0.632$\pm$0.022 & 0.596$\pm$0.003 & 0.564$\pm$0.012 \\
GPT-4o-mini & 0.769$\pm$0.004 & 0.795$\pm$0.001 & 0.225$\pm$0.003 & 0.511$\pm$0.018 & 0.207$\pm$0.005 & 0.234$\pm$0.003 \\
GPT-4.1-nano & 0.779$\pm$0.008 & 0.785$\pm$0.010 & 0.672$\pm$0.005 & 0.627$\pm$0.018 & 0.562$\pm$0.003 & 0.556$\pm$0.004 \\
Phi-4-mini & 0.587$\pm$0.005 & 0.658$\pm$0.001 & 0.248$\pm$0.004 & 0.404$\pm$0.003 & 0.421$\pm$0.006 & 0.387$\pm$0.001 \\
Llama-3.2-3B & 0.385$\pm$0.014 & 0.335$\pm$0.012 & 0.151$\pm$0.004 & 0.100$\pm$0.003 & 0.093$\pm$0.004 & 0.108$\pm$0.008 \\
Llama-3.1-8B & 0.799$\pm$0.004 & 0.833$\pm$0.001 & 0.333$\pm$0.010 & 0.667$\pm$0.012 & 0.203$\pm$0.009 & 0.383$\pm$0.001 \\
Llama-3.3-70B & 0.777$\pm$0.002 & 0.788$\pm$0.005 & 0.656$\pm$0.004 & 0.733$\pm$0.002 & 0.582$\pm$0.002 & 0.628$\pm$0.003 \\
Mistral-7B & 0.718$\pm$0.006 & 0.713$\pm$0.006 & 0.793$\pm$0.003 & 0.806$\pm$0.006 & 0.810$\pm$0.003 & 0.813$\pm$0.001 \\
Mistral-Large & 0.902$\pm$0.001 & 0.911$\pm$0.007 & 0.635$\pm$0.006 & 0.692$\pm$0.005 & 0.681$\pm$0.004 & 0.698$\pm$0.001 \\
Qwen3-8B & 0.801$\pm$0.008 & 0.841$\pm$0.001 & 0.508$\pm$0.002 & 0.624$\pm$0.006 & 0.488$\pm$0.010 & 0.514$\pm$0.000 \\
Qwen3-30B-A3B & 0.884$\pm$0.002 & 0.899$\pm$0.000 & 0.573$\pm$0.000 & 0.742$\pm$0.006 & 0.583$\pm$0.002 & 0.624$\pm$0.003 \\
Qwen3-32B & 0.876$\pm$0.003 & 0.862$\pm$0.007 & 0.730$\pm$0.009 & 0.728$\pm$0.007 & 0.545$\pm$0.002 & 0.558$\pm$0.001 \\
GLM-4.5-Air & 0.879$\pm$0.004 & 0.885$\pm$0.003 & 0.660$\pm$0.007 & 0.662$\pm$0.006 & 0.575$\pm$0.004 & 0.567$\pm$0.007 \\
\bottomrule
\end{tabular}
\end{table*}

\subsection{Reasoning Budget Sweep}
\label{app:reasoning_budget}

Reasoning models expose a control over how much internal reasoning they perform before answering. To check whether this lever changes benchmark conclusions, we sweep each provider's control over its low, medium, and high settings on EpiQAL-B and EpiQAL-C, using the \texttt{reasoning\_effort} parameter for the OpenAI models and the closest provider-specific control available for DeepSeek. Table~\ref{tab:reasoning_budget} reports the two extremes in Exact Match and average reasoning tokens. DeepSeek-V4-Pro is outside the benchmark model set and is included only as an exploratory comparison. The accuracies here come from re-running the API models and differ slightly from Table~\ref{tab:bench}, which we attribute to stochasticity and API variation.

\begin{table}[h]
\centering
\small
\setlength{\tabcolsep}{4pt}
\caption{Reasoning budget sweep, low to high setting (zero-shot, no-CoT). EM is Exact Match; tok is average reasoning tokens per question, in thousands.}
\label{tab:reasoning_budget}
\begin{tabular}{lcccc}
\toprule
\textbf{Model} & \textbf{B EM} & \textbf{B tok} & \textbf{C EM} & \textbf{C tok} \\
\midrule
GPT-5-mini & .55--.56 & 0.2--2.1 & .56--.64 & 0.3--2.5 \\
DeepSeek-V4-Flash & .70--.73 & 1.0--1.1 & .66--.66 & 1.2--1.2 \\
DeepSeek-V4-Pro & .79--.80 & 1.6--1.6 & .67--.66 & 2.1--2.1 \\
\bottomrule
\end{tabular}
\end{table}

The effect is small and model-dependent. GPT-5-mini's reasoning budget responds strongly to the setting, expanding roughly tenfold from low to high, while the DeepSeek models barely change their budget at all. This indicates that provider-specific reasoning controls are not directly comparable across families, so the sweep should not be read as a controlled comparison of reasoning depth. Accuracy changes are small on EpiQAL-B for all three models. GPT-5-mini shows a suggestive 0.08 increase on EpiQAL-C from low to high, which we do not interpret as a firm effect without a paired confidence interval. This lever is also distinct from Chain-of-Thought prompting, which unlocks explicit reasoning in models that do not reason by default, whereas the models swept here already do.

\subsection{Token Usage}
\label{app:token_usage}

We logged per-question output and reasoning tokens for the four API models. Table~\ref{tab:token_usage} averages over EpiQAL-A, B, and C; per-subset figures are cited in the text where relevant. Input length is approximately 7k tokens for every model, since all receive the same article, so differences arise on the output side. The rerun accuracies are close to those in Table~\ref{tab:bench}, with small deviations that we attribute to stochasticity and API variation.

\begin{table}[h]
\centering
\small
\caption{Average token usage per question and mean Exact Match across the three subsets (zero-shot, no-CoT).}
\label{tab:token_usage}
\begin{tabular}{lccc}
\toprule
\textbf{Model} & \textbf{Out. tok} & \textbf{Reason. tok} & \textbf{Mean EM} \\
\midrule
DeepSeek-V4 & 870 & 861 & 0.769 \\
GPT-5-mini & 632 & 615 & 0.667 \\
GPT-4o-mini & 13 & 0 & 0.402 \\
GPT-4.1-nano & 10 & 0 & 0.673 \\
\bottomrule
\end{tabular}
\end{table}

The most direct comparison is between DeepSeek-V4-Flash-Thinking and GPT-5-mini. DeepSeek reaches a higher mean Exact Match, 0.769 against 0.667, while using roughly 1.4 times as many reasoning tokens on average and roughly 1.5 times as many on EpiQAL-B, so its accuracy advantage comes with a larger reasoning budget. Separately, the non-reasoning models emit far fewer output tokens, and GPT-4.1-nano reaches a mean Exact Match close to GPT-5-mini while producing around ten output tokens per question. We report these figures descriptively rather than as a cost-efficiency comparison, since input, output, and reasoning tokens are priced differently across providers and averaging over the three subsets hides substantial differences in per-subset rankings.

\subsection{Out-of-Distribution Evaluation}
\label{app:ood_analysis}

To assess potential data contamination, we collect 96 research articles from PLOS Neglected Tropical Diseases published in March 2026, after the release dates of all evaluated models except DeepSeek-V4-Flash-Thinking. We run the full EpiQAL construction pipeline with identical settings and evaluate the 14 models under zero-shot no-CoT. DeepSeek-V4-Flash-Thinking is retained because it replaced DeepSeek-V3.2-Thinking in the API during appendix experiments. Table~\ref{tab:ood} reports Exact Match on the original and OOD sets.

\begin{table*}[htbp]
\centering
\small
\caption{Out-of-distribution evaluation: Exact Match on original vs.\ post-release articles (zero-shot, no-CoT). \textbf{Bold} = best, \underline{underline} = second best per column. Ties share the same mark.}
\label{tab:ood}
\begin{tabular}{lcccccc}
\toprule
& \multicolumn{2}{c}{\textbf{EpiQAL-A}} & \multicolumn{2}{c}{\textbf{EpiQAL-B}} & \multicolumn{2}{c}{\textbf{EpiQAL-C}} \\
\cmidrule(lr){2-3} \cmidrule(lr){4-5} \cmidrule(lr){6-7}
\textbf{Model} & Original & OOD & Original & OOD & Original & OOD \\
\midrule
DeepSeek-V4-Flash-Thinking$^\ddagger$ & \textbf{0.926} & \textbf{0.920} & 0.703 & \textbf{0.782} & 0.666 & 0.688 \\
GPT-5-mini & \underline{0.905} & 0.898 & 0.533 & 0.552 & 0.599 & 0.581 \\
GPT-4o-mini & 0.766 & 0.795 & 0.222 & 0.345 & 0.213 & 0.226 \\
GPT-4.1-nano & 0.768 & 0.795 & 0.678 & 0.678 & 0.559 & 0.516 \\
Phi-4-mini-instruct & 0.583 & 0.591 & 0.249 & 0.276 & 0.426 & 0.430 \\
Llama-3.2-3B-Instruct & 0.366 & 0.455 & 0.157 & 0.195 & 0.088 & 0.118 \\
Llama-3.1-8B-Instruct & 0.798 & 0.807 & 0.318 & 0.230 & 0.190 & 0.280 \\
Llama-3.3-70B-Instruct & 0.779 & 0.705 & 0.651 & 0.632 & 0.580 & 0.591 \\
Mistral-7B-Instruct-v0.3 & 0.722 & 0.670 & \textbf{0.789} & \textbf{0.782} & \textbf{0.808} & \textbf{0.806} \\
Mistral-Large-Instruct-2411 & 0.901 & 0.875 & 0.644 & 0.586 & \underline{0.685} & \underline{0.720} \\
Qwen3-8B & 0.811 & 0.864 & 0.508 & 0.471 & 0.484 & 0.430 \\
Qwen3-30B-A3B-Instruct-2507 & 0.882 & 0.875 & 0.573 & 0.667 & 0.585 & 0.538 \\
Qwen3-32B & 0.872 & 0.852 & \underline{0.743} & \underline{0.724} & 0.547 & 0.484 \\
GLM-4.5-Air & 0.874 & \underline{0.909} & 0.657 & \underline{0.724} & 0.580 & 0.559 \\
\bottomrule
\end{tabular}
\end{table*}

Spearman rank correlation between original and OOD model rankings is 0.94 (EpiQAL-A), 0.95 (EpiQAL-B), and 0.96 (EpiQAL-C), indicating that model performance generalizes to unseen documents and is not driven by memorization of training-set passages.

\subsection{Cross-Source Generalization}
\label{app:cross_source}

To test whether the pipeline generalizes beyond PLOS NTD, we apply it to 82 articles from the International Journal of Epidemiology (Oxford University Press), which covers broader epidemiological topics including chronic and environmental epidemiology. We use the same generation model (Qwen3-30B-A3B) and construction settings, yielding 82, 79, and 78 instances for A, B, and C respectively. Table~\ref{tab:ije} reports Exact Match for the 14 models under zero-shot no-CoT.

\begin{table*}[htbp]
\centering
\small
\caption{Cross-source evaluation: Exact Match on IJE articles (zero-shot, no-CoT). PLOS results are in Table~\ref{tab:bench}. \textbf{Bold} = best, \underline{underline} = second best per column. Ties share the same mark.}
\label{tab:ije}
\begin{tabular}{lccc}
\toprule
\textbf{Model} & \textbf{EpiQAL-A} & \textbf{EpiQAL-B} & \textbf{EpiQAL-C} \\
\midrule
DeepSeek-V4-Flash-Thinking$^\ddagger$ & \textbf{0.939} & \textbf{0.886} & \underline{0.744} \\
GPT-5-mini & 0.902 & 0.658 & 0.551 \\
GPT-4o-mini & 0.780 & 0.582 & 0.295 \\
GPT-4.1-nano & 0.805 & 0.646 & 0.526 \\
Phi-4-mini-instruct & 0.622 & 0.557 & 0.500 \\
Llama-3.2-3B-Instruct & 0.402 & 0.038 & 0.103 \\
Llama-3.1-8B-Instruct & 0.866 & 0.722 & 0.321 \\
Llama-3.3-70B-Instruct & 0.866 & 0.785 & 0.731 \\
Mistral-7B-Instruct-v0.3 & 0.756 & \underline{0.810} & \textbf{0.808} \\
Mistral-Large-Instruct-2411 & \underline{0.927} & 0.734 & 0.603 \\
Qwen3-8B & 0.854 & 0.671 & 0.526 \\
Qwen3-30B-A3B-Instruct-2507 & 0.902 & 0.759 & 0.603 \\
Qwen3-32B & \underline{0.927} & 0.759 & 0.538 \\
GLM-4.5-Air & 0.902 & 0.684 & 0.628 \\
\bottomrule
\end{tabular}
\end{table*}

Performance patterns are consistent across sources: EpiQAL-A remains the easiest subset, while B and C pose greater challenges. Key findings from the PLOS experiments replicate on IJE: Mistral-7B leads on EpiQAL-C (0.808), Llama-3.2-3B collapses on reasoning-intensive subsets, and scale does not predict success. These results suggest that the construction pipeline ports to epidemiological literature beyond neglected tropical diseases. Since we use the same generator and checkers with no separate expert pass, this is a portability check rather than evidence that the taxonomy covers epidemiology as a whole.

\subsection{Cross-Generator Robustness}
\label{app:generator_ablation}

All main-experiment instances are produced by a single generator, which raises the question of whether the observed model rankings and the near-saturation of EpiQAL-A are specific to that generator. To probe this, we regenerated a 50-article sample of each subset from the same PLOS NTD corpus using DeepSeek-V4-Pro, a stronger model from a different family than Qwen3-30B-A3B, and re-evaluated the models under zero-shot no-CoT. Table~\ref{tab:generator_ablation} compares mean Exact Match against Table~\ref{tab:bench}, which covers the full benchmark, and reports the per-model rank correlation over the 13 models that have a one-to-one counterpart in Table~\ref{tab:bench}.

\begin{table}[h]
\centering
\small
\caption{Cross-generator comparison. Mean Exact Match over models and per-model Spearman rank correlation between the two generators (zero-shot, no-CoT).}
\label{tab:generator_ablation}
\begin{tabular}{lccc}
\toprule
\textbf{Subset} & \textbf{Qwen mean} & \textbf{DeepSeek mean} & \textbf{Spearman} \\
\midrule
A & 0.77 & 0.88 & 0.86 \\
B & 0.52 & 0.40 & 0.81 \\
C & 0.49 & 0.37 & 0.74 \\
\bottomrule
\end{tabular}
\end{table}

Per-model ordering is largely preserved across generators. The 13 models compared already exclude the DeepSeek family, and dropping the Qwen family as well leaves the correlations at 0.92, 0.76, and 1.00 on A, B, and C, so the agreement is not an artifact of same-family models tracking their own generator. EpiQAL-A stays near saturation under the stronger generator, with mean Exact Match if anything higher, which is consistent with the saturation reflecting the recall-oriented nature of the task rather than a property of the original generator. EpiQAL-B and EpiQAL-C remain far from saturated and are somewhat harder under the new generator.

This analysis uses one alternative generator on a 50-article sample per subset, whereas Table~\ref{tab:bench} evaluates the full benchmark, and it does not include a matched comparison of distractor deception rates. It is therefore a preliminary robustness check rather than a controlled generator ablation, and we draw no conclusions beyond what the sample supports.

\subsection{Construction and Evaluation Roles}
\label{app:role_overlap}

Several evaluated models also take part in benchmark construction. Table~\ref{tab:roles} lists each role so the overlap can be checked directly.

\begin{table}[h]
\centering
\small
\caption{Model roles in construction and evaluation. $^{*}$ marks models whose provider family also appears in the evaluation set under a different checkpoint. GPT-5-mini additionally performs the post-hoc distractor classification in \S\ref{sec:error}, which is an analysis step rather than part of construction.}
\label{tab:roles}
\begin{tabular}{llc}
\toprule
\textbf{Model} & \textbf{Construction role} & \textbf{Evaluated} \\
\midrule
Qwen3-30B-A3B$^{*}$ & generator & yes \\
GPT-5-mini$^{*}$ & checker, diff judge & yes \\
DeepSeek-V3.2$^{*}$ & checker, diff judge & yes \\
GLM-4.5-Air & checker & yes \\
Qwen3-32B$^{*}$ & diff judge & yes \\
Phi-4-mini & diff judge & yes \\
the other nine models & none & yes \\
\bottomrule
\end{tabular}
\end{table}

No model generates the items it is later evaluated on beyond the generator itself, which ranks ninth on EpiQAL-B. Rank alone, however, does not test for generator-style bias. Beyond exact-model overlap, several evaluated models share a provider family with construction models even where the checkpoints differ, so the checker and difficulty-judge pool may share blind spots with part of the evaluation set. A fully held-out generator and judge split would be cleaner and remains a direction for future work.

\subsection{Model Version Comparison}
\label{app:model_version}

DeepSeek-V4-Flash-Thinking, a lighter variant that replaced V3.2-Thinking via API update, provides a natural comparison point. On EpiQAL-A, V4-Flash-Thinking performs nearly identically to V3.2-Thinking (0.926 vs.\ 0.928 EM without CoT). However, on the reasoning-intensive subsets, V4-Flash-Thinking drops substantially: 0.703 vs.\ 0.818 on EpiQAL-B and 0.666 vs.\ 0.720 on EpiQAL-C. This pattern is consistent with partially distinct model behaviors across retrieval and multi-step reasoning, and suggests that the lighter variant disproportionately loses reasoning performance while largely retaining retrieval accuracy.

\subsection{Detailed Discussion}
\label{app:discussion_detail}

This section provides supplementary analysis beyond the main discussion in \S\ref{sec:discussion}.

\vskip 0.2em \noindent\textbf{Per-model distractor vulnerability.}
On EpiQAL-A, DeepSeek-V3.2-Thinking achieves the lowest overall deception rate (4.7\%), while Phi-4-mini (23.8\%) and Llama-3.2-3B (21.7\%) show uniformly high rates. Across all models, Semantic near-miss is consistently the most deceptive category, with mid-tier models such as GPT-4o-mini (21.5\% on this category) and Qwen3-8B (19.0\%) showing rates roughly double their averages on other categories.

On EpiQAL-B, Mistral-7B (5.9\%) and DeepSeek-V3.2-Thinking (6.1\%) lead, while GPT-4o-mini (34.4\%) surpasses even Llama-3.2-3B (31.0\%), indicating that reasoning-level distractors expose weaknesses in models that perform adequately on retrieval.

On EpiQAL-C, Llama-3.1-8B shows the highest vulnerability (40.9\%), driven by Speculation/limitation distractors (43.4\%), while Mistral-7B maintains the lowest rate (5.5\%) with a flat profile across categories.

\vskip 0.2em \noindent\textbf{Residual weaknesses of top models.}
Even top models show category-specific residual weaknesses. DeepSeek-V3.2-Thinking reduces most EpiQAL-A categories to $\leq$2.7\% deception, yet Semantic near-miss remains at 13.6\%. On EpiQAL-C, DeepSeek-V3.2-Thinking achieves near-zero Causal reversal deception (0.7\%), suggesting that dedicated reasoning capabilities specifically aid causal chain validation under incomplete evidence.

\vskip 0.2em \noindent\textbf{GPT-5-mini ranking anomaly.}
GPT-5-mini ranks third on EpiQAL-A (EM 0.905) but drops to mid-tier on EpiQAL-B (0.634 with CoT) and EpiQAL-C (0.555 with CoT). Per-model deception analysis offers a partial explanation: its vulnerability profile is uneven across distractor categories on inference-heavy subsets, suggesting strong retrieval but weaker multi-step reasoning under domain-specific constraints.

\vskip 0.2em \noindent\textbf{Additional Chain-of-Thought effects.}
Beyond the models highlighted in the main text, CoT benefits on EpiQAL-B are widespread: Qwen3-30B-A3B improves from 0.573 to 0.747 ($+$0.174) and Llama-3.3-70B from 0.651 to 0.732 ($+$0.081). On EpiQAL-C, the mixed pattern extends further: Phi-4-mini degrades (0.426 $\to$ 0.388), while Llama-3.3-70B improves modestly (0.580 $\to$ 0.626). This suggests that CoT amplification of errors under partial evidence is not model-specific but a more general phenomenon affecting models that rely on explicit reasoning chains when the evidence base is incomplete.

\section{Prompts}
\label{app:prompt}
Tables~\ref{tab:prompt_epiqala}--\ref{tab:prompt_eval} present abbreviated generation, verification, and evaluation prompts. Full prompts are in the released code.

\begin{table*}[h!]
\caption{Prompts used for EpiQAL-A generation and verification (abbreviated; full prompts in supplementary code).}
\label{tab:prompt_epiqala}
\small
\begin{tcolorbox}
\textcolor{blue}{\textbf{Question Generation:}}
\\\\
Your task is to generate a retrieval-based question using the provided passage. The question should be answerable by directly locating information in the passage, without requiring inference or external knowledge.
\\\textit{[... ...]}\\
Target specific factual content such as numbers, dates, definitions, or clearly stated facts. The question stem should not copy phrases from the passage that would make the answer obvious, and should not be answerable by general knowledge alone. Use the topic to constrain scope but do not ask about the topic name directly.
\\\\
\textcolor{blue}{\textbf{Correct Option Generation:}}
\\\\
Correct options should be answers that can be directly found in the passage.
\\\textit{[... ...]}\\
Each option must be directly supported by explicit text. Use concise wording without copying entire evidence sentences. If generating multiple options, ensure each represents a distinct correct answer from different parts of the passage.
\\\\
\textcolor{blue}{\textbf{Distractor Generation:}}
\\\\
Distractors should be plausible-sounding answers that appear in the passage but do not correctly answer the specific question asked.
\\\textit{[... ...]}\\
Good distractors belong to the same semantic category as the correct option, are factually accurate within the passage context, but relate to a different entity, time, place, or context. Each distractor must be a valid passage fact but incorrect for this specific question. Match the style, length, and specificity of correct options.
\\\\
\textcolor{blue}{\textbf{Verification (Checker):}}
\\\\
Verify whether options are valid as \{answer\_type\} based on the passage, question, evidence, and rationale. Evaluate based on the option's role relative to the question, not on external factual accuracy.
\\\textit{[... ...]}\\
Verify evidence is an exact verbatim quote from the passage. \textit{[Correct Option]}: Verify the option directly and accurately answers the question. \textit{[Distractor]}: Focus on meaning, not surface form---reordering or rephrasing does not make an option incorrect; only a substantive factual difference counts. If any check fails, Coherence must be No.
\\
\end{tcolorbox}
\end{table*}

\begin{table*}[h!]
\caption{Prompts used for EpiQAL-B generation and verification (abbreviated; full prompts in supplementary code).}
\label{tab:prompt_epiqalb}
\small
\begin{tcolorbox}
\textcolor{blue}{\textbf{Question Generation:}}
\\\\
Your task is to generate a multiple-choice style question that requires multi-step reasoning. The question should be grounded in the passage, guided by the topic, and optionally informed by external domain knowledge.
\\\textit{[... ...]}\\
Select at least two pieces of evidence from different parts of the passage that must be combined. Establish a reasoning chain requiring an epidemiological principle, comparison, or implication not stated in any single evidence piece. Verify that the question cannot be answered from one evidence piece alone, and that the passage does not directly provide the answer. Keep the stem concise; do not embed numbers, percentages, or descriptive phrases pointing to key evidence.
\\\\
\textcolor{blue}{\textbf{Correct Option Generation:}}
\\\\
Options should be derived conclusions from integrating the provided evidence, not facts directly retrievable from the passage.
\\\textit{[... ...]}\\
Each option must require integrating at least two evidence pieces, must not be a paraphrase of any single sentence, and must require applying an epidemiological principle. Do not embed external knowledge as explicit facts in the option text. If generating multiple options, ensure they are genuinely distinct.
\\\\
\textcolor{blue}{\textbf{Distractor Generation:}}
\\\\
Distractors must look structurally identical to correct options but contain a subtle logical flaw detectable only through careful reasoning.
\\\textit{[... ...]}\\
Each distractor should target a distinct error type: confusing related concepts, applying a valid method to an incompatible study design, mixing up the target variable, using correct terminology while violating assumptions, or drawing conclusions requiring unavailable data. Verify each distractor requires domain expertise to eliminate and is definitively wrong.
\\\\
\textcolor{blue}{\textbf{Verification (Checker):}}
\\\\
Check whether options are valid as \{answer\_type\}. Verify evidence is traceable to provided materials; flag fabricated claims. Check options do not embed external-knowledge-only information and are not direct retrievals from the passage.
\\\textit{[... ...]}\\
\textit{[Correct Option]}: Can the option be logically derived by integrating multiple evidence pieces and applying epidemiological principles? The conclusion must not be reachable from any single piece alone.
\textit{[Distractor]}: Could the conclusion be reached through valid reasoning? If yes, Coherence = No. The flaw must be substantive, not just a difference in wording.
\\
\end{tcolorbox}
\end{table*}

\begin{table*}[h!]
\caption{Prompts used for EpiQAL-C generation and verification (abbreviated; full prompts in supplementary code).}
\label{tab:prompt_epiqalc}
\small
\begin{tcolorbox}
\textcolor{blue}{\textbf{Correct Option Extraction:}}
\\\\
Extract one conclusion from the Discussion section that will serve as the Correct Option. Readers will see only the Passage Body and must identify which conclusion can be derived from it.
\\\textit{[... ...]}\\
The conclusion must not be explicitly stated in the Passage Body, must be logically derivable from Passage Body evidence by applying general epidemiological principles, and must not require results from other studies, external facts, or comparisons to benchmarks not in the Passage Body. Reject direct numerical summaries from Results, study limitations, speculative statements, and generic observations. Prefer conclusions requiring integration of multiple evidence pieces.
\\\\
\textcolor{blue}{\textbf{Question Generation:}}
\\\\
Generate a question stem answerable only by the extracted Correct Option, which is a conclusion derived from the Passage Body through epidemiological reasoning.
\\\textit{[... ...]}\\
The question must require integrating multiple evidence pieces and applying epidemiological principles. The stem must not use words appearing in the Option field, must not hint at the conclusion through synonyms or paraphrases, and must not reveal which evidence pieces are relevant. Verify the question is passage-specific and not answerable by general knowledge alone.
\\\\
\textcolor{blue}{\textbf{Distractor Generation:}}
\\\\
Distractors should be plausible-sounding conclusions that cannot actually be derived from the Passage Body alone.
\\\textit{[... ...]}\\
Candidate categories: external dependency (requires information from cited studies not in Passage Body), speculation (future directions or untested hypotheses), limitations (methodological caveats), background only (general knowledge without study-specific interpretation), and causal reversal (misinterpreting cause-effect relationships from the evidence). Verify each candidate cannot answer the question through valid Passage Body reasoning. Aim for 3--4 diverse distractors.
\\\\
\textcolor{blue}{\textbf{Verification (Checker):}}
\\\\
Check whether options are valid as \{answer\_type\} based on Passage Body, Discussion, evidence, and rationale. The Passage Body contains all sections except the Discussion.
\\\textit{[... ...]}\\
Check Discussion\_Source traceability and that evidence is not fabricated. Verify the option is not a direct retrieval from the Passage Body.
\textit{[Correct Option]}: Can the conclusion be logically derived from the Passage Body by integrating evidence and applying epidemiological principles? Base this on what the Passage Body contains, not on what the generator cited.
\textit{[Distractor]}: Could this conclusion be correctly derived from the Passage Body alone? If yes, Coherence = No. Valid flaws include causal reversal, overgeneralization, requiring external information, misinterpreting findings, or presenting speculation as established conclusions.
\\
\end{tcolorbox}
\end{table*}

\begin{table*}[h!]
\caption{Evaluation prompts across all three subsets.}
\label{tab:prompt_eval}
\small
\begin{tcolorbox}
\textcolor{blue}{\textbf{EpiQAL-A (zero-shot, without CoT):}}
\\\\
You are an expert in epidemiology. Answer the retrieval-based question using the provided passage. There may be one correct answer, multiple correct answers, or no correct answer. Select all options whose answers can be directly found in the passage.
\\\\
\textcolor{blue}{\textbf{EpiQAL-A (zero-shot, with CoT):}}
\\\\
You are an expert in epidemiology. Answer the question using the provided passage. You must evaluate each option independently. For each option, locate information in the passage that is relevant to the option. Determine whether the option matches information explicitly stated in the passage and correctly answers the specific question asked.
\\\\
\textcolor{blue}{\textbf{EpiQAL-B (zero-shot, without CoT):}}
\\\\
You are an expert in epidemiology. Answer the question based on the provided passage and your epidemiological knowledge. There may be one correct answer, multiple correct answers, or no correct answer. Select all correct answers. Some options may require combining passage evidence with epidemiological principles.
\\\\
\textcolor{blue}{\textbf{EpiQAL-B (zero-shot, with CoT):}}
\\\\
You are an expert in epidemiology. Answer the question based on the provided passage and your epidemiological knowledge. You must evaluate each option independently. Some options require multi-step reasoning that combines evidence from the passage with epidemiological principles. A correct option must be supported by passage evidence and represent valid epidemiological reasoning. An incorrect option may contain methodological errors, misinterpret the evidence, or require information not available in the passage.
\\\\
\textcolor{blue}{\textbf{EpiQAL-C (zero-shot, without CoT):}}
\\\\
You are an expert in epidemiology. Answer the question using the provided Passage Body, which contains all sections of the paper except the Discussion. There may be one correct answer, multiple correct answers, or no correct answer. Select all conclusions that can be logically derived from the Passage Body.
\\\\
\textcolor{blue}{\textbf{EpiQAL-C (zero-shot, with CoT):}}
\\\\
You are an expert in epidemiology. Answer the question using the provided Passage Body, which contains all sections of the paper except the Discussion. You must evaluate each option independently. For each option, determine whether it can be logically derived from Passage Body evidence by applying general epidemiological principles. A correct option must be derivable from the Passage Body but should not be explicitly stated there verbatim. An incorrect option either requires external information not in the Passage Body, makes unsupported speculative claims, or misinterprets the evidence.
\\\\
\textit{All prompts instruct the model to output a JSON object with key ``results'' containing a list of index strings for correct options, or an empty list if none.}
\\
\end{tcolorbox}
\end{table*}

\end{document}